\documentclass{article}

\usepackage{microtype}
\usepackage{graphicx}
\usepackage{subfigure}
\usepackage{booktabs} %

\usepackage{hyperref}

\usepackage[accepted]{icml2023}
\usepackage{amsmath}
\usepackage{amssymb}
\usepackage{mathtools}
\usepackage{amsthm, bm}

\usepackage[capitalize,noabbrev]{cleveref}

\theoremstyle{plain}

\theoremstyle{definition}

\theoremstyle{remark}

\usepackage[colorinlistoftodos,textwidth=3.0cm,textsize=tiny]{todonotes}

\newcommand{\acrolong}{\textbf{C}lassification with \textbf{Hi}erarchical \textbf{L}abel \textbf{S}ets}
\newcommand{\acro}{CHiLS}
\definecolor{chng}{rgb}{0, 0., 0.}
\newcommand{\change}[1]{\textcolor{chng}{#1}}

\def\vx{{\bm{x}}}
\def\vy{{\bm{y}}}

\DeclareMathOperator*{\argmax}{arg\,max}
\usepackage{array}
\usepackage{multirow}
\usepackage{booktabs}
\usepackage{caption}
\usepackage{float}
\usepackage{longtable}
\usepackage{graphicx}
\definecolor{good}{rgb}{0.11, 0.77, 0.11}
\definecolor{bad}{rgb}{0.77, 0.11, 0.11}

\RequirePackage{algorithm}
\RequirePackage{algorithmic}
\icmltitlerunning{CHiLS: Zero-Shot Image Classification with Hierarchical Label Sets}

\begin{document}

\twocolumn[
\icmltitle{CHiLS: Zero-Shot Image Classification with Hierarchical Label Sets}

\begin{icmlauthorlist}
\icmlauthor{Zachary Novack}{ucsd}
\icmlauthor{Julian McAuley}{ucsd}
\icmlauthor{Zachary Lipton}{cmu}
\icmlauthor{Saurabh Garg}{cmu}
\end{icmlauthorlist}

\icmlaffiliation{ucsd}{Department of Computer Science and Engineering, University of California - San Diego}
\icmlaffiliation{cmu}{Machine Learning Department, Carnegie Mellon University}

\icmlcorrespondingauthor{Zachary Novack}{znovack@ucsd.edu}
\icmlcorrespondingauthor{Saurabh Garg}{sgarg2@andrew.cmu.edu}

\icmlkeywords{Zero-shot classification, }

\vskip 0.3in
]

\printAffiliationsAndNotice{}  %

\begin{abstract}
Open vocabulary models (e.g. CLIP) 
have shown strong performance on zero-shot classification 
through their ability generate embeddings for each class 
based on their (natural language) names. 
Prior work has focused on improving the accuracy of these models 
through prompt engineering or by incorporating
a small amount of labeled downstream data (via finetuning). 
\change{However, there has been little focus on improving
the richness of the class names themselves, which can pose issues
when class labels are coarsely-defined and are uninformative.}
\change{We propose \acrolong{}  (or \acro{}),
an alternative strategy
for zero-shot classification
specifically designed for datasets
with implicit semantic hierarchies.
\acro{}
proceeds in three steps}:
(i) for each class, produce a set of subclasses,
using either existing label hierarchies or by querying GPT-3; 
(ii) perform the standard zero-shot CLIP procedure 
as though these subclasses were the labels of interest;
(iii) map the predicted subclass back to its parent
to produce the final prediction. 
Across numerous datasets
with underlying hierarchical structure, 
\acro{} leads to improved accuracy
in situations 
\change{both with and without ground-truth
hierarchical information.}
\acro{} is simple to implement within existing zero-shot pipelines
and requires no additional training cost. 
Code is available at: \url{https://github.com/acmi-lab/CHILS}.

\end{abstract}

\section{Introduction}
\change{There has been a recent growth of interest in}
the capabilities of pretrained \emph{open vocabulary models} \citep{radford2021learning, wortsman2021robust, jia2021scaling, gao2021clip, pham2021scaling, Cho2022CLIPReward, rosanne2022cupl}.
These models, e.g., CLIP \citep{radford2021learning} 
and ALIGN \citep{jia2021scaling},
learn to map images and captions
into shared embedding spaces
such that images are close in embedding space
to their corresponding captions 
but far from randomly sampled captions.
The resulting models can then be used to assess 
the relative compatibility of a given image
with an arbitrary set of textual ``prompts". 
\citet{radford2021learning} observed
that by inserting each class name 
directly within a natural language prompt,
one can then use CLIP embeddings 
to perform
zero-shot image classification
with high success rates
\citep{radford2021learning, zhang2021pointclip}.

Despite the documented successes,
the current interest in open vocabulary
models poses a new question:
\textbf{How should we represent our classes 
for a given problem in natural language?}
As class names are now part of
\change{the predictive}
pipeline 
(as opposed to mostly an afterthought 
in traditional scenarios) 
for models like CLIP
in the zero-shot setting,
CLIP's performance is now directly tied
to the descriptiveness of the class ``prompts" 
\citep{santurkar2022clip}.
\change{While there is a growing body of work on improving the quality}
 of the prompts into which class names are embedded
\citep{radford2021learning, rosanne2022cupl, zhou2022coop, zhou2022cocoop, huang2022upl}, 
surprisingly little attention has been paid to
improving the \emph{richness of the class names themselves}.
This can be particularly crucial in cases 
where datasets may contain a rich underlying
structure but have uninformative class labels.
Consider, for an example, the 
\change{class}
``large man-made outdoor things" 
in the CIFAR20 dataset \citep{cifar100},
\change{which includes ``bridges" and ``roads" but
also ``castles" and ``skyscrapers"} 
(see Section \ref{sec:mot} for a more in-depth analysis).

\begin{figure*}[t!]
    \centering
    \includegraphics[width=0.9\textwidth]{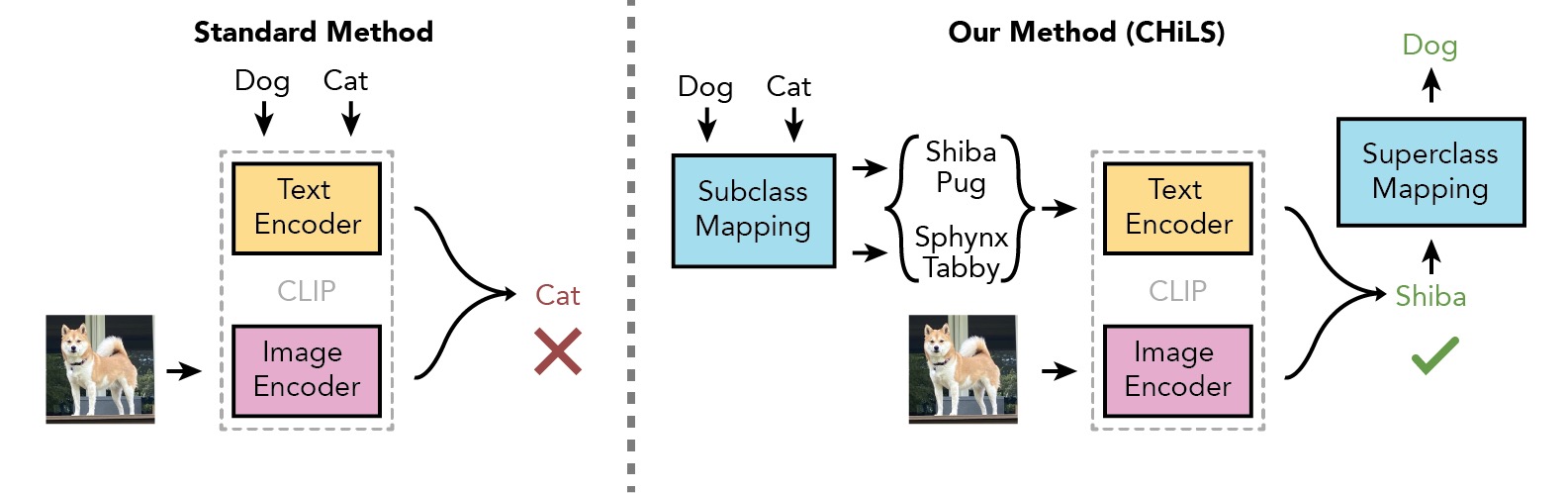}
    \caption{\textbf{(Left)} \emph{Standard CLIP Pipeline for Zero-Shot Classification}. 
    For inference, a standard CLIP takes in input a set of classes 
    and an image where we want to make a prediction 
    and makes a prediction from that set of classes. 
    \textbf{(Right)} \emph{Our proposed method CHiLS
    for leveraging hierarchical class information into the zero-shot pipeline}.
    We map each individual class to a set of subclasses, 
    perform inference in the subclass space 
    (i.e., union set of all subclasses), 
    and map the predicted subclass 
    back to its original superclass.}   
    \label{fig:mod}
\end{figure*}

\begin{figure*}
    \centering
    \includegraphics[width=0.9\textwidth]{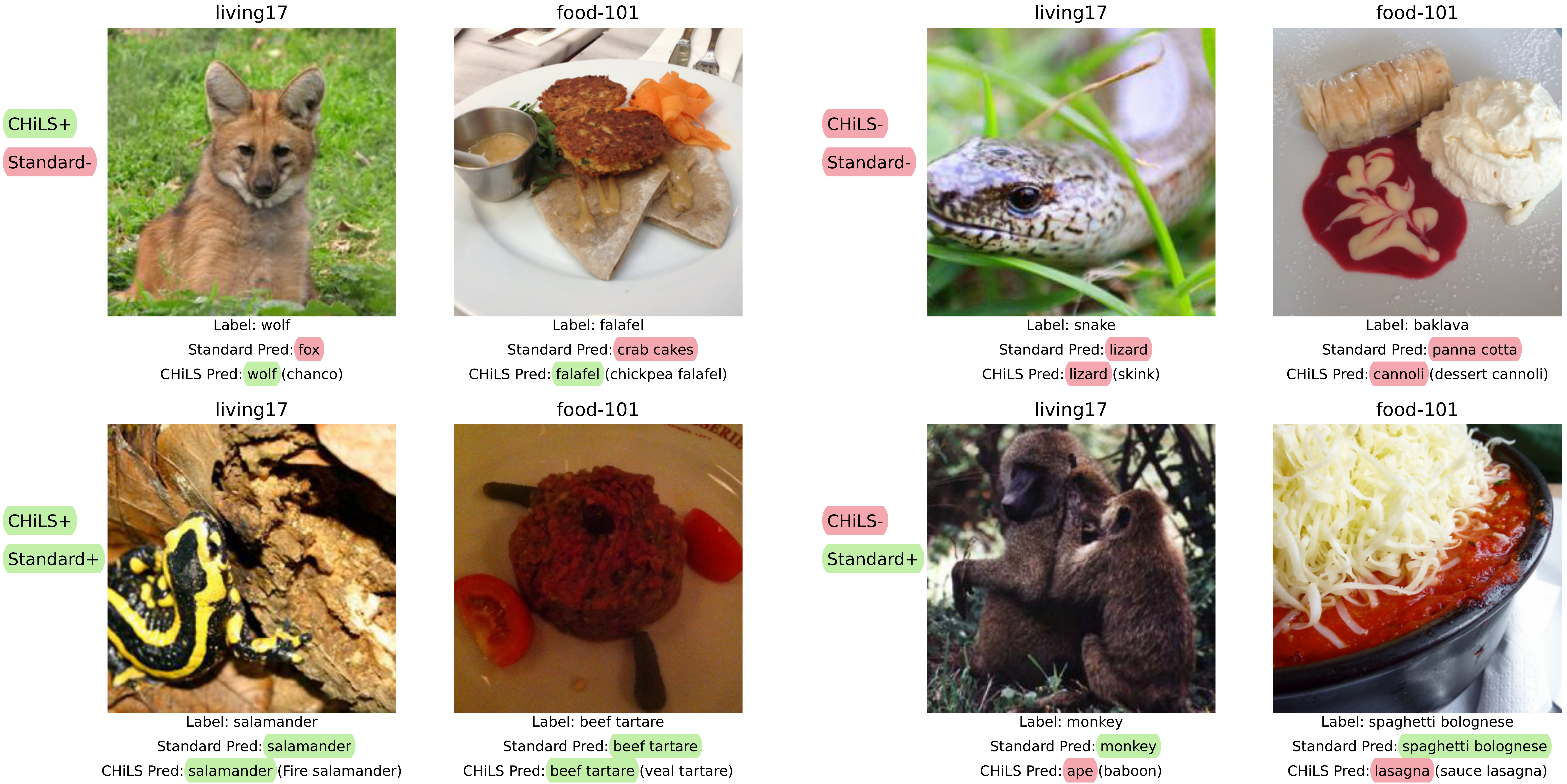}
    \caption{\change{\emph{Selected examples of behavior differences between the standard and \acro{} performance across two different datasets}. (Upper left): \acro{} is correct, standard prediction is not. (Lower left): Both correct. (Upper right): Both wrong. (Lower Right): standard prediction is correct, \acro{} is not. }} 
    \label{fig:examples}
\end{figure*}

\change{In this paper, we introduce a new method
to tackle 
zero-shot classification 
with CLIP models
for classification tasks with coarsely-defined class labels. 
We refer to our method as \acrolong{} (\acro{} for short). }
Our method utilizes a hierarchical map 
to convert each class into a list of subclasses, 
performs standard CLIP zero-shot prediction 
across the union set of all \emph{subclasses}, 
and finally uses the inverse mapping to
convert the subclass prediction to the requisite superclass. 
We additionally include a reweighting step 
wherein we leverage the raw superclass probabilities 
in order to make our method robust
to less-confident predictions at the 
superclass and subclass level.

We evaluate \acro{} on a wide array of image classification benchmarks
with \emph{and} without available hierarchical information. 
These datasets share the property of having an underlying
semantic substructure that is not captured in the initial set of class label names.
In the former case, leveraging preexisting hierarchies
leads to strong accuracy gains across all datasets. 
In the latter, we show that rather than enumerating the hierarchy by hand,
using GPT-3 to query a list of \emph{possible} subclasses for each class 
(whether or not they are actually present in the dataset) 
still leads to consistent improved accuracy over raw superclass prediction. 
We summarize our main contributions below:

\begin{itemize}
    \item We propose \acro{}, a new method for improving zero-shot CLIP performance
    \change{in scenarios with ill-defined and/or overly coarse class structures} (see Section \ref{sec:mot}), 
    which \change{only} requires 
    \change{the class names themselves}
    and is flexible to both existing 
    and synthetically generated hierarchies.
    
    \item We show that \acro{} consistently performs 
    as well or better than standard zero-shot practices
    in situations with only synthetic hierarchies, 
    and that \acro{} can achieve up to $30\%$ accuracy gains 
    when ground truth hierarchies are available. %
\end{itemize}

\section{Related Work}\label{sec:related}

\subsection{Few-Shot Learning with CLIP}

While the focus of this paper is to
improve CLIP models in the zero-shot
regime, there is a large body of work
exploring improvements to CLIP's
{few-shot} capabilities. In the standard fine-tuning
paradigm for CLIP models, practitioners 
discard the text encoder and only use
the image embeddings as inputs for some additional
training layers. 

\change{One particular line of work on improving the fine-tuned
capabilities of CLIP models leverages model weight interpolation. \citet{wortsman2021robust} propose to
linear interpolate the weights of a fine-tuned
and a zero-shot CLIP model to improve
the fine-tuned model under distribution shifts. 
This idea is extended by \citet{modelsoup} into
a general purpose paradigm for ensembling
models' weights in order to improve robustness. 
\citet{ilharco2022patching} then build on both these works
and put forth a method to ``patch" fine-tuned and zero-shot
CLIP weights together in order to avoid the issue of catastrophic forgetting. Among all the works in this section, our paper is perhaps most similar to this vein of work (albeit in spirit), as \acro{} too
seeks to combine two different predictive methods.
\citet{ding2022conti} also tackle catastrophic forgetting, 
though they propose an orthogonal direction} and
fine-tune both the image encoder and the text encoder, 
where the latter draws from a replay vocabulary of
text concepts from the original CLIP database.

\change{There is another line of work that seeks to improve CLIP
models by injecting a small amount of learnable parameters into
the frozen CLIP backbone.}
This has been commonly achieved through the adapter framework
\citep{houlsby2019parameter} from parameter-efficient learning;
specifically, in \citet{gao2021clip} they fine-tune a small
number of additional weights on top of the encoder blocks,
which is then connected with the original embeddings
through residual connections. \citet{zhang2021tip}
build on this method by removing the need for additional
training and simply uses a cached model. \change{In contrast to these
works, \citet{jia2022visual} forgo the adapter framework
when using a Vision Transformer backbone
for inserting learnable ``prompt" vectors into the 
transformer's input layers, which shows superior
performance over the aforementioned methods.}

Additionally, some have looked at circumventing
the entire process of prompt engineering. \citet{zhou2022cocoop}
and \citet{zhou2022coop} tackle this
by treating the tokens within each prompt
as learnable vectors, which are then optimized
within only a few images per class. \citet{huang2022upl} echo
these works, but instead do not utilize any labeled data
and learns the prompt representations in an unsupervised manner.
\change{\citet{zhai2022lit} completely forgo the notion of
fine-tuning in the first place, instead proposing to reframe
the pre-training process as only training a language model to match
a pre-trained and frozen image model.}
In all the above situations, \emph{some} amount of data,
whether labeled or not, is used in order to improve
the predictive accuracy of the CLIP model.

\subsection{Zero-Shot Prediction}

The field of zero-shot learning %
has existed well before the emergence
of open vocabularly models, with its
inception traced to \citet{larochelle2008zero}.
With regards to non-CLIP related methods,
the zero-shot learning paradigm has shown success in improving
multilingual question answering \citep{kuo2022qa} 
with large language models, and also in image classification
tasks where wikipedia-like context is used
in order to perform the classification
without access to the training labels 
\citep{bujwid2021large, shen2022klite}.

With CLIP models, zero-shot learning success has been found 
in a variety of tasks. Namely, \citet{zhang2021pointclip}
expand the CLIP 2D paradigm for 3D point clouds.
\citet{tewel2021i2t} show that CLIP models
can be retrofitted to perform the reverse task
of image-to-text generation, \change{and \citet{shen2021much}
likewise display CLIP's ability to improve performance
on an array of Vision\&Language tasks.}
Both \citet{yu2022esper} and \citet{Cho2022CLIPReward}
expand CLIP's zero-shot abilities through techniques
drawn from reinforcement learning (RL), with the former
using CLIP for the task of audio captioning. 
\change{\citet{gadre2022clip} similarly work with
the RL literature and retrofit CLIP to
improve the embodied AI task of object navigation without
any additional training.}
\citet{zeng2022socraticmodels} show 
the capabilities of composing CLIP-like
models and LLMs together
to extend the zero-shot capabilities
to tasks like assitive dialogue
and open-ended reasoning. Unlike our work
here, these prior directions mostly focus
on generative problems or, in the case of 
\citet{bujwid2021large} and \citet{shen2022klite},
require rich external knowledge databases to
employ their methods.

In the realm of improving CLIP's zero-shot
capabilities for image classification, 
we particularly note the contemporary work of 
\citet{rosanne2022cupl}.
Here, authors explore
using GPT-3 to generate rich textual prompts
for each class rather than using preexisting
prompt templates, and show improvements in  
zero-shot accuracy across a variety of image
classification baselines. 
In another work, 
\citet{ren2022open}
propose leveraging preexisting captions
in order to improve performance,
though this is restricted to querying
the pre-training set of captions.
In contrast, our work explores a 
complementary direction of  
leveraging hierarchy in 
class names to improve zero-shot performance of CLIP  
with a fixed set of preexisting prompt templates.

\subsection{\change{Hierarchical Classification}}
Our work is
related to Hierarchical Classification \citep{Silla2010ASO},
i.e., classification tasks when the set of labels can
be arranged in a DAG-like class hierarchy. Methodologies from
hierarchical classification have been extensively used for multi-label classification (\citet{DIMITROVSKI20112436, liu2021improving}, and \citet{chalkidis2020empirical} to name a few), and recent works have shown that this paradigm
can aid in zero-shot learning by attempting to uncover hierarchical
relations between classes \citep{chen2021hsva, mensink2014costa} and/or leveraging
existing hierarchical information during training \citep{yi2022exploring,CAO2020107488}.
While our work is similar in spirit to prior work on hierarchical classification, we note that there are two crucial distinctions:
(i) we
are concerned only with the zero-shot
\emph{training-free} regime (as we only
require class names during inference) while most previous
work assumes some amount of training, and (ii) 
\acro{} only leverages the class hierarchy for 
the flat task of \emph{superclass} prediction
without requiring  
any supervision at the subclass level.

\section{Proposed Method: \acro{}}

In this paper, we are primarily concerned with
the problem of 
zero-shot image classification in CLIP models
\change{(see App. \ref{app:primer} for an introduction to CLIP and relevant terminology)}. 
For CLIP models,
zero-shot classification
involves using both a pretrained
image encoder and a
pretrained text encoder
(see the left part of Figure \ref{fig:mod}).
To perform zero-shot classification, 
we need a predefined set of classes 
written in natural language. 
Let $\mathcal{C} = \{c_1, c_2, \dots, c_k\}$ 
be such a set.  
Given an image and set of classes,
each class is embedded within a natural
language prompt (through some function $\text{T}(\cdot)$) to produce a ``caption"
for each class (e.g. one standard prompt
mentioned in \citet{radford2021learning} is 
\emph{\change{``A photo of a \{\}."}}).
These prompts are then fed into the text encoder and
after passing the image through the image encoder,
we calculate the cosine similarity between the
image embedding and each class-prompt embedding.
These similarity scores form the output ``logits"
of the CLIP model, which can be passed through
a softmax to generate the class probabilities.

While prior works have
focused on improving
the $\text{T}(\cdot)$ for each class label
$c_i$ (refer to Section~\ref{sec:related}), we instead focus on the complementary task
of directly modifying the set of classes $\mathcal{C}$
\change{when $\mathcal{C}$ is ill-formed or overly general},
keeping $\text{T}(\cdot)$ fixed.
Our method involves two main steps:
\change{(1) performing zero-shot prediction over
label \emph{subclasses}}
and 
\change{(2) aligning subclass probabilities
with the raw superclass outputs to reconcile
both inference methods.} Next, we describe our proposed method. 

\subsection{Zero-Shot Prediction with Hierarchical Label Sets}

Our method \acro{} slightly modifies
the standard approach for zero-shot
CLIP prediction.
As each class label $c_i$ represents
some concept in natural language
(e.g. the label ``dog"),
we 
acquire a %
\textbf{subclass set} $\mathcal{S}_{c_i} = \{s_{c_i,1}, s_{c_i,2}, \dots, s_{c_i,m_i}\}$ through some mapping function $G$,
where each $s_{c_i,j}$ is a linguistic
\emph{hyponym}, or subclass, of $c_i$ 
(e.g. corgi for dogs) and
$m_i$ is the size of the set $S_{c_i}$.

Given a label set $S_{c_i}$ for each class,
we proceed with the standard process
for zero-shot prediction, but now using
the \emph{union} of all label sets as the
set of classes.
\change{Through this, \acro{} will output
a distribution over all subclasses $\widehat{\vy}_{\text{sub}}$.}
We then leverage the inverse mapping function $G^{-1}$
to 
\change{map}
\change{the argmax subclass probability} back into
the corresponding superclass \change{$G^{-1}(\argmax \widehat\vy_{\text{sub}})$}. 
Our method is detailed more formally
in Algorithm \ref{alg:zshotpred}.

In our work, we experiment with 
two scenarios: 
(i) when hierarchy information is available and can be
readily queried; 
and (ii) when hierarchy information is \emph{not} 
available and the label set for each
class must be generated, which we do so
by prompting GPT-3.

\setlength{\textfloatsep}{12pt}
\begin{algorithm*}[t]
    \caption{ \acrolong{} (\acro{})}\label{alg:zshotpred}
    \begin{algorithmic}[1]
    \INPUT:  data point $\vx$, class labels $\mathcal{C}$, prompt function $\text{T}$, label set mapping $G$, CLIP model $f$
        \vspace{3pt}
        \STATE Set $\mathcal{C}_{\text{sub}} \gets \cup_{c_i \in \mathcal{C}}^{} G(c_i)$
        \hfill \textcolor{blue}{\COMMENT{\footnotesize\texttt{Union of subclasses for subclass prediction}}} 
        \vspace{3pt}
        \STATE $\widehat{\vy}_{\text{sub}} = \sigma(f(\vx, \text{T}(\mathcal{C}_{\text{sub}})))$ 
        \hfill \textcolor{blue}{\COMMENT{\footnotesize\texttt{Subclass probabilities}} }
        \vspace{3pt}
        \STATE $\widehat\vy_{\text{sup}} = \sigma(f(\vx, \text{T}(\mathcal{C})))$
        \hfill \textcolor{blue}{\COMMENT{\footnotesize\texttt{Superclass probabilities}} }
        \vspace{3pt}
        \FOR{$i=1$ to $|\mathcal{C}|$}
            \STATE $S_{c_i} = G(c_i)$
            \FOR{$s_{c_i, j} \in S_{c_i}$}
                \STATE $\widehat{\vy}_{\text{sub}}[s_{c_i, j}] = \widehat{\vy}_{\text{sub}}[s_{c_i, j}] * \widehat{\vy}_{\text{sup}}[c_i]$ \\[2pt]
                \hfill \textcolor{blue}{\COMMENT{\footnotesize\texttt{Combining subclass and superclass prediction probability}} }
            \ENDFOR
        \ENDFOR
        \vspace{3pt}
    \OUTPUT: $G^{-1}(\argmax \widehat\vy_{\text{sub}})$
\end{algorithmic}
\end{algorithm*}

\subsection{Reweighting Probabilities With Superclass Confidence}

While the above method is able to effectively
utilize CLIP's ability to identify 
relatively fine-grained concepts, 
by predicting on only subclass labels
we lose any positive benefits of the superclass
label, and performance may vary widely based
on the quality of the subclass labels.
Given recent evidence \citep{minderer2021revisiting,kadavath2022know} that
large language models (like the text encoder in
CLIP) \change{are well-calibrated and} generally 
\change{assign higher probability to}
correct \change{predictions,}
we modify our initial algorithm
to leverage this behavior and 
\change{use}
\emph{both} superclass and subclass information.
We provide empirical evidence of this property in 
Appendix \ref{app:eecc}.

Specifically, we include an 
additional reweighting step
within our main algorithm
(see lines 4-9 in Algorithm \ref{alg:zshotpred}).
Here, we reweight each set of subclass probabilities by its
superclass probability.
\change{Heuristically, as the prediction is now taken as the argmax
over \emph{products} of probabilities, 
large disagreements between subclass and
superclass probabilities will be down-weighted (especially
if one particular superclass is confident) and subclass
probabilities will be more important in cases where the
superclass probabilities are roughly uniform.
We show ablations
on the choice of the reweighting algorithm in Section \ref{sec:abl}.}

\section{A Motivating Example for CHiLS}\label{sec:mot}
Before validating the effectiveness of
\acro{} across standard benchmarks,
we provide a more nuanced investigation
on the ImageNet dataset at different hierarchy levels. 
Given that ImageNet is 
arranged in a rich taxonomical structure,
we perform zero-shot classification at
progressively finer levels
of the hierarchy, where \acro{} is given
access to all the leaf nodes in each class
at the current level (unless the classes are themselves
leaf nodes).

In Table \ref{tab:motexp}, we see that at lower depths (e.g. depth 1 or 2), \acro{}
significantly improves on top of standard zero-shot performance. 
As the depth in the hierarchy increase, the gap between \acro{}'s performance 
and the standard zero shot decreases while the number of leaf nodes increases.
This behavior highlights a key fact about \acro's potential use cases: \emph{\acro{} can help for tasks
where class labels resemble intermediate nodes of the ImageNet hierarchy.} 

\begin{table}[ht!]
    \caption{\emph{Zero-Shot performance at different levels of ImageNet hierarchy}, where \acro{} has access to true ImageNet leaf node classes. \acro{} shows clear performance gains over the baseline at coarse-to-intermediate granularities.}
    \label{tab:motexp}
    \centering
    \begin{small}
        \begin{tabular}{cccc}
        \toprule
        ImageNet Depth & Standard & \acro{} & \% Leaf Classes\\
        \midrule
         1 & 67.43 & 97.08 & 0.0\\
         2 & 69.22 & 90.47 & 0.0\\
         3 & 63.97 & 86.20 & 0.0\\
         4 & 49.48 & 80.31 & 32.03\\
         5 & 63.80 & 74.08 & 77.90\\
         6 & 62.96 & 65.07 & 96.28\\
         \bottomrule
    \end{tabular}
    \end{small}
\end{table}

\section{Experiments}

\begin{table}[ht!]
    \caption{\emph{Zero-shot accuracy performance across 16 image benchmarks with superclass labels (baseline)}, \acro{} with existing hierarchy (whenever available), and \acro{} with GPT-3 generated hierarchy. \acro{} improves classification accuracy in all situations with given label sets and all but 2 datasets with GPT-3 generated label sets.}\label{tab:mainres}
    \vskip 0.0004in
    \begin{center}
        \begin{small}
            \begin{tabular}{lccc}
    \toprule
    \multirow{2}{*}{Dataset} & \multirow{2}{*}{Superclass} & \acro{}  & \acro{}  \\
     & &  (True Map) &  (GPT-3 Map)\\
    \midrule
    Nonliving26 & 79.8  & 90.7  \textcolor{good}{($+10.9 $)} & 81.7  \textcolor{good}{($+1.9 $)}\\
    Living17 & 91.1 & 93.8 \textcolor{good}{($+2.7 $)} & 91.6 \textcolor{good}{($+0.5 $)}\\
    Entity13 & 77.5  & 92.6  \textcolor{good}{($+15.1 $)} & 78.1  \textcolor{good}{($+0.7 $)}\\
    Entity30 & 70.3  & 88.9  \textcolor{good}{($+18.5 $)} & 71.7   \textcolor{good}{($+1.4 $)}\\
    CIFAR20 & 59.6  & 85.3  \textcolor{good}{($+25.7 $)} & 65.0 \textcolor{good}{($+5.4 $)} \\
    Food-101 & \change{93.9} & N/A & \change{93.8}  \textcolor{bad}{($-0.1$)}\\
    Fruits-360 & \change{58.8}  & \change{59.2}  \textcolor{good}{($+0.5 $)} & \change{60.1}  \textcolor{good}{($+1.4 $)} \\ 
    Fashion1M & \change{45.8}  & N/A & \change{47.4}  \textcolor{good}{($+1.7 $)}\\
    Fashion- & \multirow{2}{*}{68.5}  & \multirow{2}{*}{N/A} & \multirow{2}{*}{70.8 \textcolor{good}{($+2.2 $)}}\\
    MNIST & & & \\
    LSUN-Scene & 88.1  & N/A & 88.8  \textcolor{good}{($+0.7 $)}\\
    Office31 & \change{89.1}  & N/A & \change{90.5}  \textcolor{good}{($+1.4 $)}\\
    OfficeHome & 88.8  & N/A & 88.8  \textcolor{bad}{($-0.0$)}\\
    ObjectNet & 53.1  & 85.3  \textcolor{good}{($+32.2 $)} & 53.5  \textcolor{good}{($+0.4 $)} \\
    \change{EuroSAT} & \change{62.1} & \change{N/A} &\change{62.4} \textcolor{good}{($+0.3$)} \\
    \change{RESISC45} & \change{72.6} & \change{N/A} & \change{72.7} \textcolor{good}{($+0.1$)}\\
    \bottomrule
    \end{tabular}
        \end{small}
    \end{center}
    \vskip -0.1in
\end{table}

\subsection{Setup}\label{sec:setup}

\paragraph{Datasets.} \change{As we are primarily concerned with improving zero-shot CLIP
performance in situations with uninformative and/or semantically coarse class labels as described in Section \ref{sec:mot},}
we test our method on the 16 following image benchmarks:
the four BREEDS imagenet subsets
(Living17, Nonliving26, Entity13, and Entity30) \citep{santurkar2020breeds},
CIFAR20 (the coarse-label version of CIFAR100; ~\citet{cifar100}),
Food-101 \citep{food101}, Fruits-360 \citep{fruits360}, Fashion1M \citep{fashion1M}, Fashion-MNIST \citep{fashionmnist},
LSUN-Scene \citep{lsun}, Office31 \citep{office31}, OfficeHome \citep{officehome}, 
ObjectNet \citep{objectnet}, \change{EuroSAT \citep{eurosat1,eurosat2}, and RESISC45 \citep{resisc45}}.
We use the validation sets for each dataset (if present).
These datasets constitute a wide range of 
different image domains and include datasets
with and without available hierarchy information.
\change{Additionally, the chosen datasets vary
widely in the semantic granularity of their classes,
from overly general cases (CIFAR20) to settings with
a mixture of general and specific classes (Food-101, OfficeHome).}

We also examine \acro{}'s robustness to
distribution shift within a dataset by averaging
all results for the BREEDS datasets, Office31,
and OfficeHome across different shifts
(see Appendix \ref{app:dataset} for more information).
We additionally modify the Fruits-360 and ObjectNet
datasets to create existing taxonomies.
More details for dataset preparation 
are detailed in Appendix \ref{app:dataset}. %

\paragraph{Model Architecture.} 
Unless otherwise specified, we use the
ViTL/14@336px backbone \citep{radford2021learning}
for our CLIP model, and used DaVinci-002
(with temperature fixed at $0.7$) for
all ablations involving GPT-3. 
For the choice of the prompt embedding function
$\text{T} (\cdot)$,
\change{for each dataset we 
experiment (where applicable) with two different functions: (1)  Using the average text embeddings of the 75
    different prompts for each label used for ImageNet in
    \citet{radford2021learning}, where the prompts
    cover a wide array of captions and 
    (2) Following the procedure that \citet{radford2021learning}
    puts forth for more specialized datasets, we modify the standard prompt
    to be of the form \emph{``A photo of a \{\}, a type of \textbf{[context]}.''},
    where \textbf{[context]} is dataset-dependent
    (e.g. ``food" in the case of food-101).
In the case that a custom prompt set exists for a dataset,
as is the case with multiple datasets that the present
work shares with \citet{radford2021learning}, we use
the given prompt set for the latter option rather than building
it from scratch. For each dataset, we use the prompt set
that gives us the best \emph{baseline} (i.e. superclass) zero-shot
performance. More details are in Appendix \ref{app:context}.
}

\paragraph{Choice of Mapping Function $G$.}\label{par:map} In our experiments,
we primarily look at 
how the choice of the mapping function $G$
influences the performance of \acro{}. 
In Section \ref{sec:ghier}, we focus on the datasets
with available hierarchy information. Here, $G$ and $G^{-1}$ are
simply table lookups to find
the list of subclasses and corresponding
superclass respectively. In Section \ref{sec:gunk},
we explore situations in which the true set of subclasses
in each superclass is unknown. In these scenarios,
we use GPT-3 to generate our mapping
function $G$. Specifically, given some label set size $m$, 
superclass name \texttt{class-name},
\change{and optional \textbf{context} (which we use 
whenever using the context-based prompt embedding),}
we query GPT-3 with the prompt:

\begin{center}
    \texttt{Generate a list of \textbf{m} types of the following \change{\textbf{[context]}}: \textbf{class-name}}
\end{center}

The resulting output list from GPT-3 thus defines
our mapping $G$ from superclass to subclass.
Unless otherwise specified, we fix $m=10$ for
all datasets.
Note here that $m$ is only fixed for \emph{GPT-generated} sets, as the true label sets may have variable sizes for each superclass in a given dataset.
Additionally, in Section \ref{par:imgn} we explore situations in which
hierarchical information is present 
but noisy, i.e. the label set
for each superclass contains
\change{the true subclasses \emph{and} erroneous
subclasses that are not present in the dataset.}

\subsection{Leveraging Available Hierarchy Information}\label{sec:ghier}

We first concern ourselves with the scenario
where hierarchy information is already
available for a given 
dataset. In this situation, the set of subclasses
for each superclass is specified and correct 
(i.e. every image within each superclass falls into one
of the subclasses). We emphasize that here we do \emph{not} need 
information about which example belongs to which subclass, we just need a 
mapping of superclass to subclass. 
For example, each class in the BREEDS
dataset living17 is made up of $4$--$8$ ImageNet subclasses
at finer granularity (e.g. `parrot' includes
`african grey' and `macaw').

\paragraph{Results.} In Table \ref{tab:mainres}, 
we can see that our method performs better
than using the baseline superclass labels alone across
all 7 of the datasets with available hierarchy information,
often leading to +15\% improvements in accuracy.

\subsection{\acro{} in Unknown Hierarchy Settings}\label{sec:gunk}

Though we have seen considerable success in situations
with access to the true hierarchical structure, 
in some real-world settings our dataset %
may not include any available information about 
the subclasses within each class.
In this scenario, we turn to using GPT-3
to approximate the hierarchical map $G$
(as specified in Section \ref{par:map}).
It is important to note that GPT-3 may
sometimes output suboptimal label sets,
most notably in situations where GPT-3 chooses
the wrong wordsense or when GPT-3 only lists modifiers
on the original superclass 
(e.g. producing the list \texttt{[red, yellow, green]} for types of apples). 
In order to account for these issues in an 
out-of-the-box fashion, we make two adjustments: (i) 
append the superclass name (if not already present)
to each generated subclass label, and (ii) include
the superclass itself within the label set.
\change{For a controlled analysis about the effect of including
the superclass itself in the label set, see Appendix \ref{app:supclass}.} 

\paragraph{Results.} In this setting, 
our method is still able to beat the baseline performance
in most datasets, albeit with lower accuracy
gains  (see Table \ref{tab:mainres}). 
Thus, while knowing the true subclass hierarchy can lead to large
accuracy gains, it is enough to simply enumerate a list of possible
subclasses for each class with no prior information about the dataset
in order to improve the predictive accuracy. 
{\change{In Figure 
\ref{fig:examples}, we show selected examples to highlight \acro{}'s behavior across two datasets.
}}

\subsection{Ablations}\label{sec:abl}

\begin{table}[t]
    \caption{Average accuracy across datasets for superclass prediction, \acro{} (ours), and \acro{} \emph{without} the reweighting step. While when given the true hierarchy omitting the reweighting step can slightly boost performance beyond \acro{}, in situations without the true hierarchy the reweighting step is crucial to improving on the baseline accuracy.}
    \label{tab:rw}
    \begin{center}
    \begin{small}
    \begin{tabular}{lc}
    \toprule
    Experiment & Average Accuracy
    \\
    \midrule
    Standard &  73.28\\
    \acro{} (True Map, No RW) & 86.40 \\
    \acro{} (True Map, RW) & 85.11 \\
    \acro{} (GPT Map, No RW) & 71.61 \\
    \acro{} (GPT Map, RW) & 74.49 \\
    \bottomrule
    \end{tabular}
    \end{small}
    \end{center}
\end{table}

\paragraph{Is Reweighting Necessary?} Though the
reweighting step in \acro{} is motivated by the
evidence that CLIP generally assigns higher probability
to \emph{correct} predictions rather than incorrect
ones (see Appendix \ref{app:eecc} for empirical verification),
it is not immediately
clear whether \change{the reweighting step}
is truly necessary.
Averaged across all documented datasets,
in \change{Table \ref{tab:rw}}  we show that
in the true hierarchy setting, not reweighting
the subclass probabilities can actually slightly
\emph{boost} performance (as the label sets are
adequately tuned to the distribution of images).
However, in situations where the true hierarchy
is not present, omitting the reweighting step
puts accuracy below the baseline performance.
\change{We attribute this difference in behavior
to the fact that reweighting multiplicatively combines
the superclass and subclass predictions, and thus
if subclass performance is sufficient on its own
(as is the case when the true hierarchy is available)
then combining it with superclass predictions can cause the model
to more closely follow the behavior of the underperforming superclass predictor.}
\change{Thus, as the presence of a ground-truth hierarchy is not
guaranteed in the wild, the reweighting step is necessary
for \acro{} to improve zero-shot performance.}

\begin{table}[t]
    \caption{Average accuracy across datasets with GPT-generated label sets for different reweighting algorithms. Using aggregate subclass probabilities for reweighting performs noticeably worse than our initial method and reweighting in superclass space. \acro{} too only performs slightly worse than the contrived best possible union of subclass and superclass predictions.}
    \label{tab:rw_typ}
    \begin{center}
        \begin{small}
            \begin{tabular}{lc}
    \toprule
    Experiment & Average Accuracy
    \\
    \midrule
    Best Possible & 78.69 \\
    Standard &  73.28\\
    \acro{} &  74.49\\
    \acro{} (RW subclass w/mean subclass) & 72.79 \\
    \acro{} (RW mean subclass w/superclass)& 74.45 \\
    \bottomrule
    \end{tabular}
        \end{small}
    \end{center}
\end{table}

\paragraph{Different Reweighting Strategies.}\change{We \change{also} experimented 
with different mechanisms for reweighting
superclass and subclass predictions. Namely, we investigated
whether superclass probabilities could be replaced
by the sum over the matching subclass probabilities, \emph{and}
whether we can aggregate subclass probabilities and reweight
them with the matching superclass probabilities
(i.e. performing the normal reweighting step but in the space of superclasses).} \change{In Table \ref{tab:rw_typ} we show that replacing the superclass
probabilities in the reweighting step with aggregate subclass probabilities
removes any accuracy gains from \acro{}, but doing the reweighting step
in superclass space \emph{does} maintain \acro{} accuracy performance. 
This suggests that the beneficial behavior of \acro{} may be due to
successfully combining two different sets of class labels. We also display the upper bound for combining superclass and subclass
prediction (i.e. the accuracy when a datum is correctly labeled if the
superclass \emph{or} subclass predictions are correct), which we note
is impossible in practice, and observe that even the best possible
performance is not much higher than the performance of \acro{}.}

\paragraph{Noisy Available Hierarchies}\label{par:imgn}  While
the situation described in Section \ref{sec:gunk} is the most
probable in practice,
we additionally investigate the situation
in which the hierarchical
information is present 
but \emph{overestimates} the set of subclasses. 
\change{For example, the scenario in which
a dataset with the class ``dog" includes huskies
and corgis, but \acro{} is provided with huskies,
corgis, \emph{and labradors} as possible subclasses, with the last
being out-of-distribution.}
To do this,
we return to the BREEDS datasets presented in \citet{santurkar2020breeds}.
As the BREEDS datasets were created 
so that each class contains
the same number of subclasses
(which are ImageNet classes),
we modify $G$ such that the label
set for each superclass corresponds
to \emph{all} the ImageNet classes descended
from that node in the hierarchy
\change{(see Appendix \ref{app:noise} for more information)}.
As we can see in Table \ref{tab:gimgnet},
\acro{} is able to improve upon the baseline
performance even in the presence of added noise
in each label set.

\begin{table}[t]
\caption{\acro{} zero-shot accuracy when $G$ includes \emph{all} subclasses in the ImageNet hierarchy descended from the respective root node. Even in the presence of noise added to the true label sets, \acro{} is provides large accuracy gains.}
    \label{tab:gimgnet}
    \begin{center}
        \begin{small}
    \begin{tabular}{lccc}
    \toprule
    \multirow{2}{*}{Dataset} & \multirow{2}{*}{Standard} & \acro{} - & \acro{} - True \\
     & & True Map & Map + Noise \\
    \midrule
    nonliving26 & 79.8  & 90.7  \textcolor{good}{($+10.9 $)} & 89.8 \textcolor{good}{($+10.0$)}\\
    living17 & 91.1 & 93.8 \textcolor{good}{($+2.7 $)} & 93.2 \textcolor{good}{($+2.1$)}\\
    entity13 & 77.5  & 92.6  \textcolor{good}{($+15.1 $)} & 90.7 \textcolor{good}{($+13.2$)}\\
    entity30 & 70.3  & 88.9  \textcolor{good}{($+18.6 $)} & 86.7 \textcolor{good}{($+16.4$)}\\
    \bottomrule
    \end{tabular}
    \end{small} 
    \end{center}
\end{table}

\begin{figure}[ht!]
    \centering
    \includegraphics[width=0.5\textwidth]{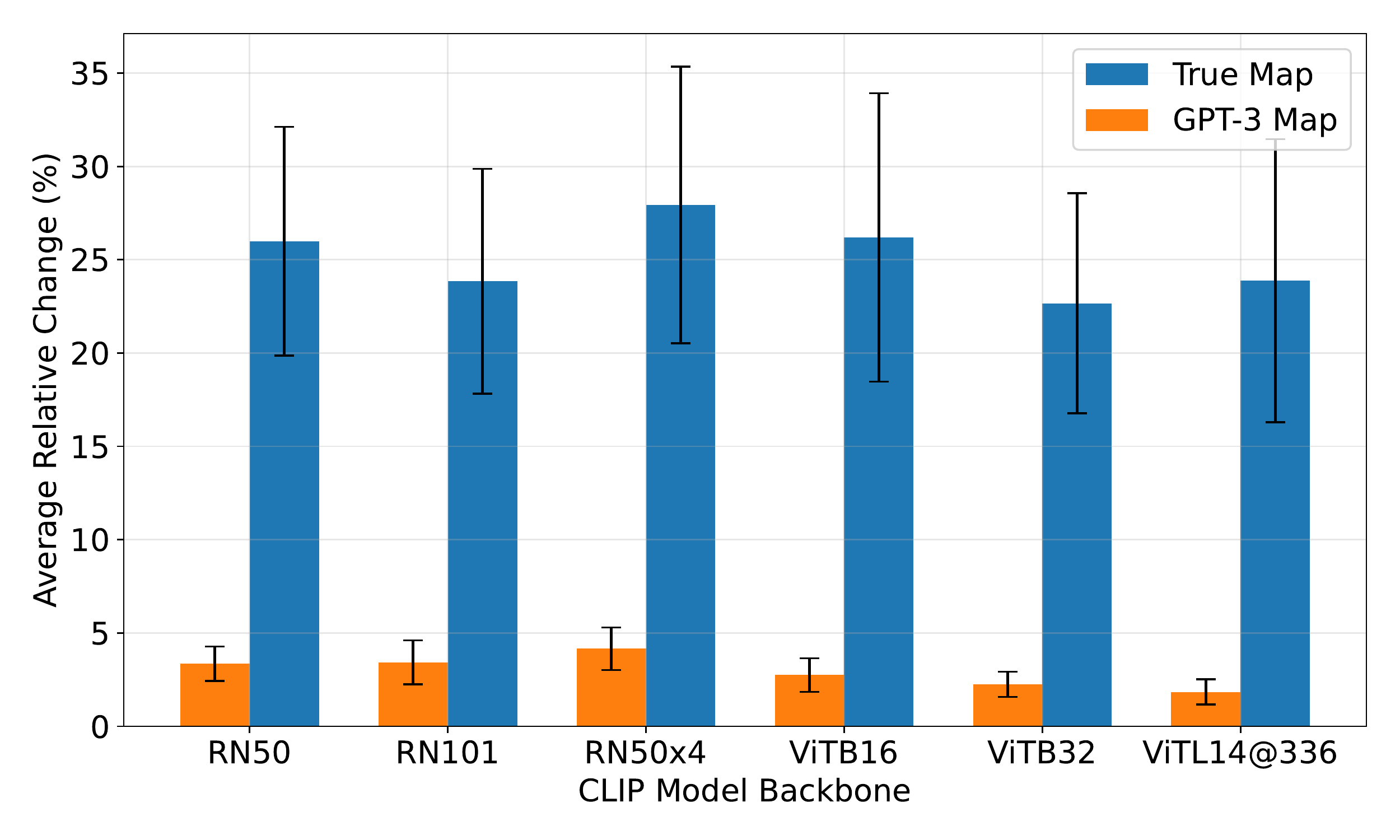}
    \includegraphics[width=0.5\textwidth]{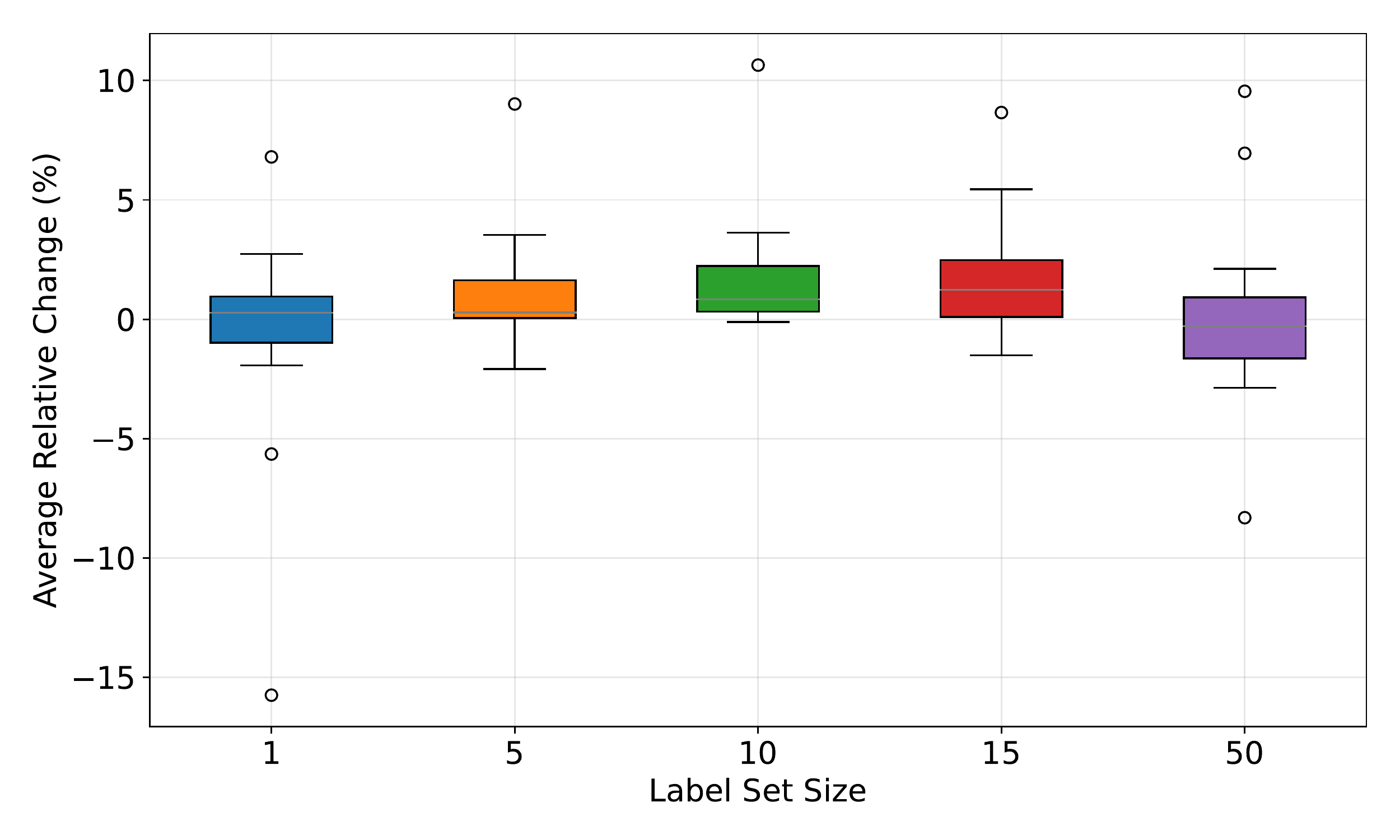}
    \caption{\change{(Top)} Average relative change between \acro{} and baseline for true mapping and GPT-3 generated mapping. Across changes in CLIP backbone size and structure, the effectiveness of \acro{} at improving performance only varies slightly. \change{(Bottom) Average relative accuracy change from the baseline to \acro{} (across all datasets), for varying label set sizes. In all, there is not much difference in performance across label set sizes.}}
    \label{fig:sizeabl}
\end{figure}

\paragraph{Label Set Size.} In previous works investigating 
importance of prompts in CLIP's performance,  
it has been documented that the number
of prompts used can have a decent effect on the overall
performance \citep{rosanne2022cupl, santurkar2022clip}.
Along this line, we investigate how the size of the
\emph{subclass set} generated for each class effects the overall
accuracy by re-running our main experiments
with varying values of $m$ (namely, \change{1}, 5, 10, 15, \change{and 50}). 
In \change{Figure \ref{fig:sizeabl}} (bottom), there is little variation
across label set sizes that is consistent over all datasets,
with the exception of the extreme label set sizes which have a few low-performing outliers.
We observe that the optimal label set size is
context-specific, and depends upon the total number of
classes present and the semantic granularity of the
classes themselves. Individual dataset results are
available in Appendix \ref{app:lsaa}.

\paragraph{Model Size.} To
examine whether the performance of
\acro{} continues to hold with 
CLIP backbones  other than ViT-L/14@336, we measure
the average relative change in accuracy performance
between \acro{} and the baseline superclass predictions
across all datasets for an array of different
CLIP models. Namely, we investigate the RN50, RN101,
RN50x4, ViT-B/16, ViT-B/32, and ViT-L/14@336 CLIP backbones (see \citet{radford2021learning} for more information
on the model specifications). In Figure \ref{fig:sizeabl} \change{(top)},
we show that across the 6 specified CLIP backbones,
\acro{} performance leads to relatively consistent
relative accuracy gains, \change{with a slight 
(but not confidently significant) trend showing
improved performance for the ResNet backbones over
the ViT backbones, which is to be expected
given their worse base capabilities.}
Thus, \acro{}'s
benefits do not seem to be 
an artifact of model scaling.

\paragraph{Alternative Aggregating Methods.} 
\change{We experimented with alternative aggregation 
methods for different parts of the \acro{} pipeline, though we found that the proposed design
(i.e. using a set-based mapping for aggregating subclasses
together and linear averaging for aggregating prompt templates)
performed the best (see Appendix \ref{app:aer} for more).}

\section{Conclusion}

In this work, we demonstrated
that the zero-shot
image classification capabilities
of CLIP models can be improved
by leveraging 
hierarchical information for a
given set of classes. 
When hierarchical structure is available
in a given dataset, our method shows
large improvements in zero-shot accuracy,
and even when subclass information \emph{isn't} 
explicitly present,
we showed that we can leverage GPT-3
to generate subclasses for each class 
and still improve upon the baseline 
(superclass) accuracy.

We remark that
\acro{} may be quite beneficial to practitioners using
CLIP as an out-of-the-box image classifier. Namely, 
we show that 
\change{in scenarios where the class
labels may be ill-formed or overly coarse,}
even without existing hierarchical data
accuracy can be improved with a \emph{fully automated}
pipeline (via querying GPT-3), yet \acro{} is flexible
enough that any degree of hand-crafting label sets can
be worked into the zero-shot pipeline. Our method 
has the added benefit of being both
\emph{completely zero-shot} (i.e. no training
or fine-tuning necessary) and is resource 
efficient.

\paragraph{Limitations and Future Work.}
As with usual zero-shot learning, 
we don't have a way to validate the performance 
of our method. 
\change{Additionally, we recognize that \acro{}
is suited for scenarios in which a semantic hierarchy
likely
exists, and thus may not be particularly useful
in classification tasks where the classes are already
fine-grained.
We believe that this limitation will not hinder 
the applicability of our method,
as practitioners would know if their task 
contains any latent semantic hierarchy and thus choose to use our method or not \emph{a priori}.}
Given \acro{}'s empirical successes,
we hope to perform more investigation 
to develop an understanding of \emph{why}
\acro{} is able to improve zero-shot accuracy
and whether there is a more principled way of 
reconciling superclass and subclass predictions.

\subsection*{Acknowledgments}

SG acknowledges Amazon Graduate Fellowship and  JP Morgan AI Ph.D. Fellowship for their support. 
ZL acknowledges Amazon AI, Salesforce Research, Facebook, UPMC, Abridge, the PwC Center, the Block Center, the Center for Machine Learning and Health, and the CMU Software Engineering Institute (SEI) via Department of Defense contract FA8702-15-D-0002. ZN acknowledges Aaron Broukhim for their help in figure design, as well as Honey the Shiba Inu for donating their likeness to the opening figure.

\bibliography{paper}

\begin{thebibliography}{55}
\providecommand{\natexlab}[1]{#1}
\providecommand{\url}[1]{\texttt{#1}}
\expandafter\ifx\csname urlstyle\endcsname\relax
  \providecommand{\doi}[1]{doi: #1}\else
  \providecommand{\doi}{doi: \begingroup \urlstyle{rm}\Url}\fi

\bibitem[Barbu et~al.(2019)Barbu, Mayo, Alverio, Luo, Wang, Gutfreund,
  Tenenbaum, and Katz]{objectnet}
Barbu, A., Mayo, D., Alverio, J., Luo, W., Wang, C., Gutfreund, D., Tenenbaum,
  J., and Katz, B.
\newblock Objectnet: A large-scale bias-controlled dataset for pushing the
  limits of object recognition models.
\newblock In \emph{Advances in Neural Information Processing Systems
  (NeurIPS)}, 2019.

\bibitem[Bossard et~al.(2014)Bossard, Guillaumin, and Van~Gool]{food101}
Bossard, L., Guillaumin, M., and Van~Gool, L.
\newblock Food-101 -- mining discriminative components with random forests.
\newblock In \emph{European Conference on Computer Vision (ECCV)}, 2014.

\bibitem[Bujwid \& Sullivan(2021)Bujwid and Sullivan]{bujwid2021large}
Bujwid, S. and Sullivan, J.
\newblock Large-scale zero-shot image classification from rich and diverse
  textual descriptions.
\newblock In \emph{Proceedings of the Third Workshop on Beyond Vision and
  LANguage: inTEgrating Real-world kNowledge (LANTERN)}, 2021.

\bibitem[Cao et~al.(2020)Cao, Lu, Cui, and Zhang]{CAO2020107488}
Cao, Z., Lu, J., Cui, S., and Zhang, C.
\newblock Zero-shot handwritten chinese character recognition with hierarchical
  decomposition embedding.
\newblock \emph{Pattern Recognition}, 2020.

\bibitem[Chalkidis et~al.(2020)Chalkidis, Fergadiotis, Kotitsas, Malakasiotis,
  Aletras, and Androutsopoulos]{chalkidis2020empirical}
Chalkidis, I., Fergadiotis, M., Kotitsas, S., Malakasiotis, P., Aletras, N.,
  and Androutsopoulos, I.
\newblock An empirical study on large-scale multi-label text classification
  including few and zero-shot labels.
\newblock \emph{arXiv preprint arXiv:2010.01653}, 2020.

\bibitem[Chen et~al.(2021)Chen, Xie, Liu, Peng, Sun, Li, You, and
  Shao]{chen2021hsva}
Chen, S., Xie, G., Liu, Y., Peng, Q., Sun, B., Li, H., You, X., and Shao, L.
\newblock Hsva: Hierarchical semantic-visual adaptation for zero-shot learning.
\newblock 2021.

\bibitem[Cheng et~al.(2017)Cheng, Han, and Lu]{resisc45}
Cheng, G., Han, J., and Lu, X.
\newblock Remote sensing image scene classification: Benchmark and state of the
  art.
\newblock \emph{Proceedings of the IEEE}, 2017.

\bibitem[Cho et~al.(2022)Cho, Yoon, Kale, Dernoncourt, Bui, and
  Bansal]{Cho2022CLIPReward}
Cho, J., Yoon, S., Kale, A., Dernoncourt, F., Bui, T., and Bansal, M.
\newblock Fine-grained image captioning with clip reward.
\newblock In \emph{Findings of NAACL}, 2022.

\bibitem[Deng et~al.(2009)Deng, Dong, Socher, Li, Li, and
  Fei-Fei]{deng2009imagenet}
Deng, J., Dong, W., Socher, R., Li, L.-J., Li, K., and Fei-Fei, L.
\newblock {ImageNet: A Large-Scale Hierarchical Image Database}.
\newblock In \emph{Computer Vision and Pattern Recognition (CVPR)}, 2009.

\bibitem[Dimitrovski et~al.(2011)Dimitrovski, Kocev, Loskovska, and
  Džeroski]{DIMITROVSKI20112436}
Dimitrovski, I., Kocev, D., Loskovska, S., and Džeroski, S.
\newblock Hierarchical annotation of medical images.
\newblock \emph{Pattern Recognition}, 2011.

\bibitem[Ding et~al.(2022)Ding, Liu, Tian, Yang, and Ding]{ding2022conti}
Ding, Y., Liu, L., Tian, C., Yang, J., and Ding, H.
\newblock Don't stop learning: Towards continual learning for the clip model,
  2022.

\bibitem[Gadre et~al.(2022)Gadre, Wortsman, Ilharco, Schmidt, and
  Song]{gadre2022clip}
Gadre, S.~Y., Wortsman, M., Ilharco, G., Schmidt, L., and Song, S.
\newblock Clip on wheels: Zero-shot object navigation as object localization
  and exploration.
\newblock \emph{arXiv preprint arXiv:2203.10421}, 2022.

\bibitem[Gao et~al.(2021)Gao, Geng, Zhang, Ma, Fang, Zhang, Li, and
  Qiao]{gao2021clip}
Gao, P., Geng, S., Zhang, R., Ma, T., Fang, R., Zhang, Y., Li, H., and Qiao, Y.
\newblock Clip-adapter: Better vision-language models with feature adapters.
\newblock \emph{arXiv preprint arXiv:2110.04544}, 2021.

\bibitem[Gao et~al.(2020)Gao, Fisch, and Chen]{gao2020making}
Gao, T., Fisch, A., and Chen, D.
\newblock Making pre-trained language models better few-shot learners.
\newblock \emph{arXiv preprint arXiv:2012.15723}, 2020.

\bibitem[Helber et~al.(2018)Helber, Bischke, Dengel, and Borth]{eurosat2}
Helber, P., Bischke, B., Dengel, A., and Borth, D.
\newblock Introducing eurosat: A novel dataset and deep learning benchmark for
  land use and land cover classification.
\newblock 2018.

\bibitem[Helber et~al.(2019)Helber, Bischke, Dengel, and Borth]{eurosat1}
Helber, P., Bischke, B., Dengel, A., and Borth, D.
\newblock Eurosat: A novel dataset and deep learning benchmark for land use and
  land cover classification.
\newblock \emph{IEEE Journal of Selected Topics in Applied Earth Observations
  and Remote Sensing}, 2019.

\bibitem[Houlsby et~al.(2019)Houlsby, Giurgiu, Jastrzebski, Morrone,
  De~Laroussilhe, Gesmundo, Attariyan, and Gelly]{houlsby2019parameter}
Houlsby, N., Giurgiu, A., Jastrzebski, S., Morrone, B., De~Laroussilhe, Q.,
  Gesmundo, A., Attariyan, M., and Gelly, S.
\newblock Parameter-efficient transfer learning for nlp.
\newblock In \emph{International Conference on Machine Learning (ICML)}, 2019.

\bibitem[Huang et~al.(2022)Huang, Chu, and Wei]{huang2022upl}
Huang, T., Chu, J., and Wei, F.
\newblock Unsupervised prompt learning for vision-language models, 2022.

\bibitem[Ilharco et~al.(2022)Ilharco, Wortsman, Gadre, Song, Hajishirzi,
  Kornblith, Farhadi, and Schmidt]{ilharco2022patching}
Ilharco, G., Wortsman, M., Gadre, S.~Y., Song, S., Hajishirzi, H., Kornblith,
  S., Farhadi, A., and Schmidt, L.
\newblock Patching open-vocabulary models by interpolating weights.
\newblock \emph{arXiv preprint arXiv:2208.05592}, 2022.

\bibitem[Jia et~al.(2021)Jia, Yang, Xia, Chen, Parekh, Pham, Le, Sung, Li, and
  Duerig]{jia2021scaling}
Jia, C., Yang, Y., Xia, Y., Chen, Y.-T., Parekh, Z., Pham, H., Le, Q., Sung,
  Y.-H., Li, Z., and Duerig, T.
\newblock Scaling up visual and vision-language representation learning with
  noisy text supervision.
\newblock In \emph{International Conference on Machine Learning (ICML)}, 2021.

\bibitem[Jia et~al.(2022)Jia, Tang, Chen, Cardie, Belongie, Hariharan, and
  Lim]{jia2022visual}
Jia, M., Tang, L., Chen, B.-C., Cardie, C., Belongie, S., Hariharan, B., and
  Lim, S.-N.
\newblock Visual prompt tuning.
\newblock \emph{arXiv preprint arXiv:2203.12119}, 2022.

\bibitem[Kadavath et~al.(2022)Kadavath, Conerly, Askell, Henighan, Drain,
  Perez, Schiefer, Dodds, DasSarma, Tran-Johnson, Johnston, El-Showk, Jones,
  Elhage, Hume, Chen, Bai, Bowman, Fort, Ganguli, Hernandez, Jacobson, Kernion,
  Kravec, Lovitt, Ndousse, Olsson, Ringer, Amodei, Brown, Clark, Joseph, Mann,
  McCandlish, Olah, and Kaplan]{kadavath2022know}
Kadavath, S., Conerly, T., Askell, A., Henighan, T., Drain, D., Perez, E.,
  Schiefer, N., Dodds, Z.~H., DasSarma, N., Tran-Johnson, E., Johnston, S.,
  El-Showk, S., Jones, A., Elhage, N., Hume, T., Chen, A., Bai, Y., Bowman, S.,
  Fort, S., Ganguli, D., Hernandez, D., Jacobson, J., Kernion, J., Kravec, S.,
  Lovitt, L., Ndousse, K., Olsson, C., Ringer, S., Amodei, D., Brown, T.,
  Clark, J., Joseph, N., Mann, B., McCandlish, S., Olah, C., and Kaplan, J.
\newblock Language models (mostly) know what they know, 2022.

\bibitem[Krizhevsky(2009)]{cifar100}
Krizhevsky, A.
\newblock Learning multiple layers of features from tiny images.
\newblock Technical report, 2009.

\bibitem[Kuo \& Chen(2022)Kuo and Chen]{kuo2022qa}
Kuo, C.-C. and Chen, K.-Y.
\newblock Toward zero-shot and zero-resource multilingual question answering.
\newblock \emph{IEEE Access}, 2022.

\bibitem[Larochelle et~al.(2008)Larochelle, Erhan, and
  Bengio]{larochelle2008zero}
Larochelle, H., Erhan, D., and Bengio, Y.
\newblock Zero-data learning of new tasks.
\newblock In \emph{Association for the Advancement of Artificial Intelligence
  (AAAI)}, 2008.

\bibitem[Liu et~al.(2021)Liu, Zhang, Yin, and Zhu]{liu2021improving}
Liu, H., Zhang, D., Yin, B., and Zhu, X.
\newblock Improving pretrained models for zero-shot multi-label text
  classification through reinforced label hierarchy reasoning.
\newblock \emph{arXiv preprint arXiv:2104.01666}, 2021.

\bibitem[Mensink et~al.(2014)Mensink, Gavves, and Snoek]{mensink2014costa}
Mensink, T., Gavves, E., and Snoek, C.~G.
\newblock Costa: Co-occurrence statistics for zero-shot classification.
\newblock In \emph{Proceedings of the IEEE conference on computer vision and
  pattern recognition}, 2014.

\bibitem[Minderer et~al.(2021)Minderer, Djolonga, Romijnders, Hubis, Zhai,
  Houlsby, Tran, and Lucic]{minderer2021revisiting}
Minderer, M., Djolonga, J., Romijnders, R., Hubis, F., Zhai, X., Houlsby, N.,
  Tran, D., and Lucic, M.
\newblock Revisiting the calibration of modern neural networks.
\newblock In \emph{Advances in Neural Information Processing Systems
  (NeurIPS)}, 2021.

\bibitem[Mureşan \& Oltean(2018)Mureşan and Oltean]{fruits360}
Mureşan, H. and Oltean, M.
\newblock Fruit recognition from images using deep learning.
\newblock \emph{Acta Universitatis Sapientiae, Informatica}, 2018.

\bibitem[Pham et~al.(2021)Pham, Dai, Ghiasi, Kawaguchi, Liu, Yu, Yu, Chen,
  Luong, Wu, et~al.]{pham2021scaling}
Pham, H., Dai, Z., Ghiasi, G., Kawaguchi, K., Liu, H., Yu, A.~W., Yu, J., Chen,
  Y.-T., Luong, M.-T., Wu, Y., et~al.
\newblock Combined scaling for open-vocabulary image classification.
\newblock \emph{arXiv preprint arXiv: 2111.10050}, 2021.

\bibitem[Pratt et~al.(2022)Pratt, Liu, and Farhadi]{rosanne2022cupl}
Pratt, S., Liu, R., and Farhadi, A.
\newblock What does a platypus look like? generating customized prompts for
  zero-shot image classification, 2022.

\bibitem[Radford et~al.(2021)Radford, Kim, Hallacy, Ramesh, Goh, Agarwal,
  Sastry, Askell, Mishkin, Clark, et~al.]{radford2021learning}
Radford, A., Kim, J.~W., Hallacy, C., Ramesh, A., Goh, G., Agarwal, S., Sastry,
  G., Askell, A., Mishkin, P., Clark, J., et~al.
\newblock Learning transferable visual models from natural language
  supervision.
\newblock In \emph{International Conference on Machine Learning (ICML)}, 2021.

\bibitem[Ren et~al.(2022)Ren, Li, Ren, Zhao, and Sun]{ren2022open}
Ren, S., Li, L., Ren, X., Zhao, G., and Sun, X.
\newblock Rethinking the openness of clip, 2022.

\bibitem[Saenko et~al.(2010)Saenko, Kulis, Fritz, and Darrell]{office31}
Saenko, K., Kulis, B., Fritz, M., and Darrell, T.
\newblock Adapting visual category models to new domains.
\newblock In \emph{European Conference on Computer Vision (ECCV)}, 2010.

\bibitem[Santurkar et~al.(2021)Santurkar, Tsipras, and
  Madry]{santurkar2020breeds}
Santurkar, S., Tsipras, D., and Madry, A.
\newblock Breeds: Benchmarks for subpopulation shift.
\newblock In \emph{International Conference on Learning Representations
  (ICLR)}, 2021.

\bibitem[Santurkar et~al.(2022)Santurkar, Dubois, Taori, Liang, and
  Hashimoto]{santurkar2022clip}
Santurkar, S., Dubois, Y., Taori, R., Liang, P., and Hashimoto, T.
\newblock Is a caption worth a thousand images? a controlled study for
  representation learning, 2022.

\bibitem[Shen et~al.(2021)Shen, Li, Tan, Bansal, Rohrbach, Chang, Yao, and
  Keutzer]{shen2021much}
Shen, S., Li, L.~H., Tan, H., Bansal, M., Rohrbach, A., Chang, K.-W., Yao, Z.,
  and Keutzer, K.
\newblock How much can clip benefit vision-and-language tasks?
\newblock \emph{arXiv preprint arXiv:2107.06383}, 2021.

\bibitem[Shen et~al.(2022)Shen, Li, Hu, Xie, Yang, Zhang, Rohrbach, Gan, Wang,
  Yuan, Liu, Keutzer, Darrell, and Gao]{shen2022klite}
Shen, S., Li, C., Hu, X., Xie, Y., Yang, J., Zhang, P., Rohrbach, A., Gan, Z.,
  Wang, L., Yuan, L., Liu, C., Keutzer, K., Darrell, T., and Gao, J.
\newblock K-lite: Learning transferable visual models with external knowledge,
  2022.

\bibitem[Silla \& Freitas(2010)Silla and Freitas]{Silla2010ASO}
Silla, C.~N. and Freitas, A.~A.
\newblock A survey of hierarchical classification across different application
  domains.
\newblock \emph{Data Mining and Knowledge Discovery}, 2010.

\bibitem[Tewel et~al.(2021)Tewel, Shalev, Schwartz, and Wolf]{tewel2021i2t}
Tewel, Y., Shalev, Y., Schwartz, I., and Wolf, L.
\newblock Zerocap: Zero-shot image-to-text generation for visual-semantic
  arithmetic, 2021.

\bibitem[Venkateswara et~al.(2017)Venkateswara, Eusebio, Chakraborty, and
  Panchanathan]{officehome}
Venkateswara, H., Eusebio, J., Chakraborty, S., and Panchanathan, S.
\newblock Deep hashing network for unsupervised domain adaptation.
\newblock In \emph{Conference on Computer Vision and Pattern Recognition
  (CVPR)}, 2017.

\bibitem[Wortsman et~al.(2021)Wortsman, Ilharco, Kim, Li, Kornblith, Roelofs,
  Gontijo-Lopes, Hajishirzi, Farhadi, Namkoong, and
  Schmidt]{wortsman2021robust}
Wortsman, M., Ilharco, G., Kim, J.~W., Li, M., Kornblith, S., Roelofs, R.,
  Gontijo-Lopes, R., Hajishirzi, H., Farhadi, A., Namkoong, H., and Schmidt, L.
\newblock Robust fine-tuning of zero-shot models.
\newblock \emph{arXiv preprint arXiv:2109.01903}, 2021.

\bibitem[Wortsman et~al.(2022)Wortsman, Ilharco, Gadre, Roelofs, Gontijo-Lopes,
  Morcos, Namkoong, Farhadi, Carmon, Kornblith, et~al.]{modelsoup}
Wortsman, M., Ilharco, G., Gadre, S.~Y., Roelofs, R., Gontijo-Lopes, R.,
  Morcos, A.~S., Namkoong, H., Farhadi, A., Carmon, Y., Kornblith, S., et~al.
\newblock Model soups: averaging weights of multiple fine-tuned models improves
  accuracy without increasing inference time.
\newblock In \emph{International Conference on Machine Learning}, pp.\
  23965--23998. PMLR, 2022.

\bibitem[Xiao et~al.(2017)Xiao, Rasul, and Vollgraf]{fashionmnist}
Xiao, H., Rasul, K., and Vollgraf, R.
\newblock Fashion-mnist: a novel image dataset for benchmarking machine
  learning algorithms, 2017.

\bibitem[{Xiao} et~al.(2010){Xiao}, {Hays}, {Ehinger}, {Oliva}, and
  {Torralba}]{sun397}
{Xiao}, J., {Hays}, J., {Ehinger}, K.~A., {Oliva}, A., and {Torralba}, A.
\newblock Sun database: Large-scale scene recognition from abbey to zoo.
\newblock In \emph{2010 IEEE Computer Society Conference on Computer Vision and
  Pattern Recognition}, 2010.

\bibitem[Xiao et~al.(2015)Xiao, Xia, Yang, Huang, and Wang]{fashion1M}
Xiao, T., Xia, T., Yang, Y., Huang, C., and Wang, X.
\newblock Learning from massive noisy labeled data for image classification.
\newblock In \emph{Conference on Computer Vision and Pattern Recognition
  (CVPR)}, 2015.

\bibitem[Yi et~al.(2022)Yi, Shen, Gou, and Elhoseiny]{yi2022exploring}
Yi, K., Shen, X., Gou, Y., and Elhoseiny, M.
\newblock Exploring hierarchical graph representation for large-scale zero-shot
  image classification.
\newblock \emph{arXiv preprint arXiv:2203.01386}, 2022.

\bibitem[Yu et~al.(2015)Yu, Seff, Zhang, Song, Funkhouser, and Xiao]{lsun}
Yu, F., Seff, A., Zhang, Y., Song, S., Funkhouser, T., and Xiao, J.
\newblock Lsun: Construction of a large-scale image dataset using deep learning
  with humans in the loop, 2015.

\bibitem[Yu et~al.(2022)Yu, Chung, Yun, Hessel, Park, Lu, Ammanabrolu, Zellers,
  Bras, Kim, and Choi]{yu2022esper}
Yu, Y., Chung, J., Yun, H., Hessel, J., Park, J., Lu, X., Ammanabrolu, P.,
  Zellers, R., Bras, R.~L., Kim, G., and Choi, Y.
\newblock Multimodal knowledge alignment with reinforcement learning, 2022.

\bibitem[Zeng et~al.(2022)Zeng, Attarian, Ichter, Choromanski, Wong, Welker,
  Tombari, Purohit, Ryoo, Sindhwani, Lee, Vanhoucke, and
  Florence]{zeng2022socraticmodels}
Zeng, A., Attarian, M., Ichter, B., Choromanski, K., Wong, A., Welker, S.,
  Tombari, F., Purohit, A., Ryoo, M., Sindhwani, V., Lee, J., Vanhoucke, V.,
  and Florence, P.
\newblock Socratic models: Composing zero-shot multimodal reasoning with
  language.
\newblock \emph{arXiv}, 2022.

\bibitem[Zhai et~al.(2022)Zhai, Wang, Mustafa, Steiner, Keysers, Kolesnikov,
  and Beyer]{zhai2022lit}
Zhai, X., Wang, X., Mustafa, B., Steiner, A., Keysers, D., Kolesnikov, A., and
  Beyer, L.
\newblock Lit: Zero-shot transfer with locked-image text tuning.
\newblock In \emph{Proceedings of the IEEE/CVF Conference on Computer Vision
  and Pattern Recognition}, 2022.

\bibitem[Zhang et~al.(2021{\natexlab{a}})Zhang, Fang, Gao, Zhang, Li, Dai,
  Qiao, and Li]{zhang2021tip}
Zhang, R., Fang, R., Gao, P., Zhang, W., Li, K., Dai, J., Qiao, Y., and Li, H.
\newblock Tip-adapter: Training-free clip-adapter for better vision-language
  modeling.
\newblock \emph{arXiv preprint arXiv:2111.03930}, 2021{\natexlab{a}}.

\bibitem[Zhang et~al.(2021{\natexlab{b}})Zhang, Guo, Zhang, Li, Miao, Cui,
  Qiao, Gao, and Li]{zhang2021pointclip}
Zhang, R., Guo, Z., Zhang, W., Li, K., Miao, X., Cui, B., Qiao, Y., Gao, P.,
  and Li, H.
\newblock Pointclip: Point cloud understanding by clip.
\newblock \emph{arXiv preprint arXiv:2112.02413}, 2021{\natexlab{b}}.

\bibitem[Zhou et~al.(2022{\natexlab{a}})Zhou, Yang, Loy, and
  Liu]{zhou2022cocoop}
Zhou, K., Yang, J., Loy, C.~C., and Liu, Z.
\newblock Conditional prompt learning for vision-language models.
\newblock In \emph{IEEE/CVF Conference on Computer Vision and Pattern
  Recognition (CVPR)}, 2022{\natexlab{a}}.

\bibitem[Zhou et~al.(2022{\natexlab{b}})Zhou, Yang, Loy, and Liu]{zhou2022coop}
Zhou, K., Yang, J., Loy, C.~C., and Liu, Z.
\newblock Learning to prompt for vision-language models.
\newblock \emph{International Journal of Computer Vision (IJCV)},
  2022{\natexlab{b}}.

\end{thebibliography}
\bibliographystyle{icml2023}

\newpage

\onecolumn 
\appendix
\section*{Appendix}

\section{Empirical Evidence of CLIP Confidence}\label{app:eecc}

\begin{figure}[H]
     \centering
     \subfigure[nonliving26]{\label{fig:nlconf}\includegraphics[width=0.8\textwidth]{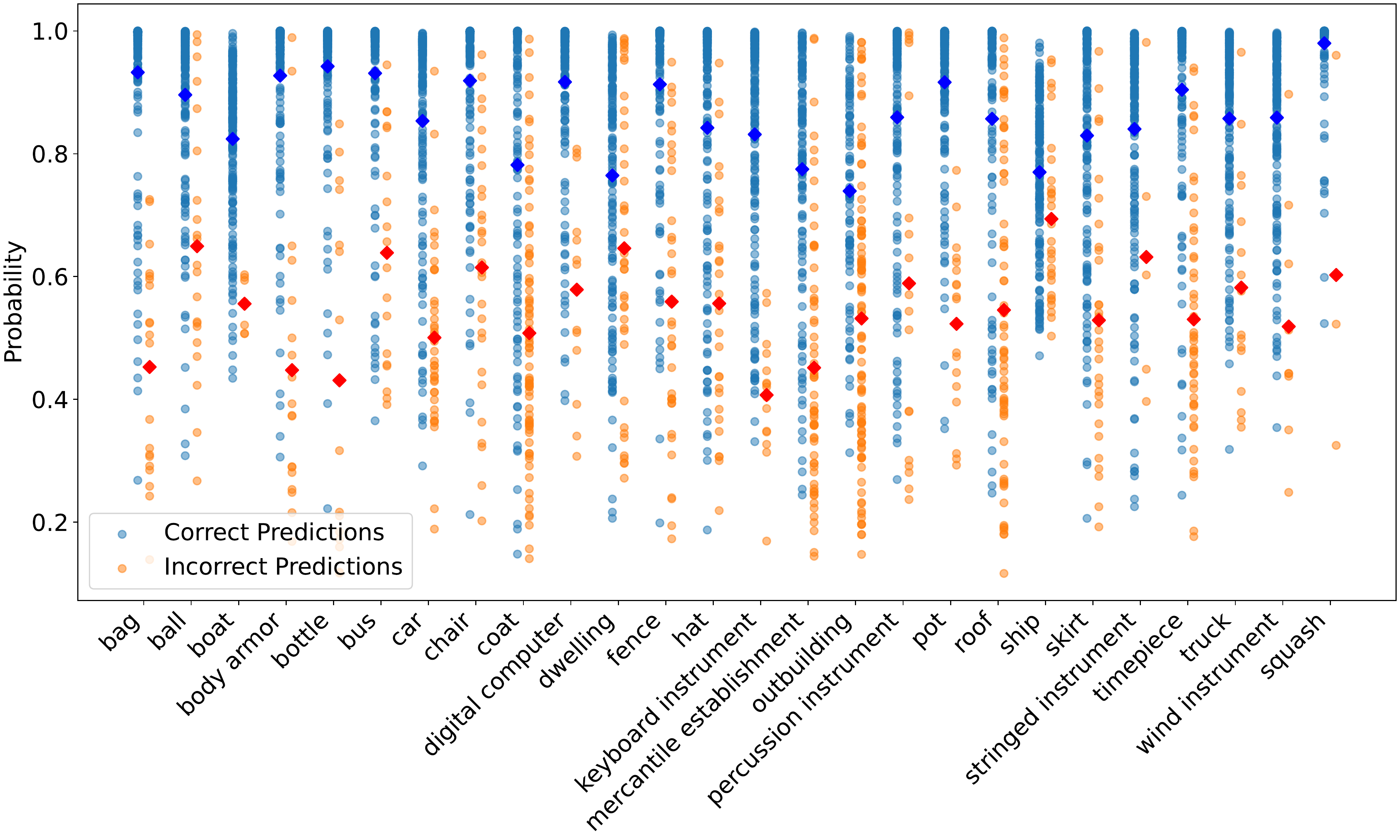}}
     \subfigure[living17]{\label{fig:lconf}\includegraphics[width=0.8\textwidth]{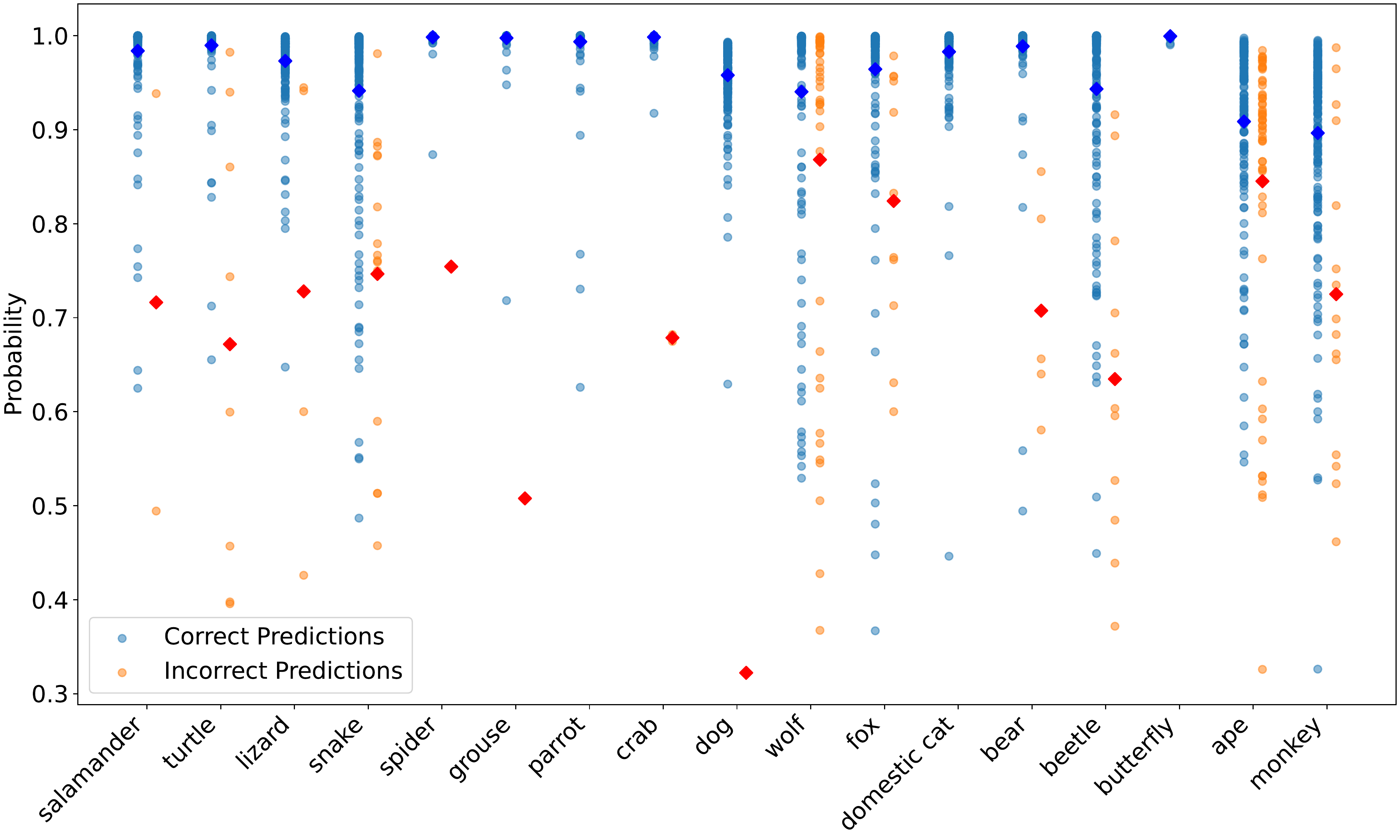}}
     
     \caption{Distribution of argmax probabilities across ImageNet BREEDS datasets for correctly and incorrectly classified data points, with the diamonds representing average probability for each class. Correctly classified probabilities are on average higher than the misclassified probabilities.}
        \label{fig:confs}
\end{figure}
\begin{figure}[H]\ContinuedFloat
    \centering
    \subfigure[entity13]{\label{fig:e13conf}\includegraphics[width=0.8\textwidth]{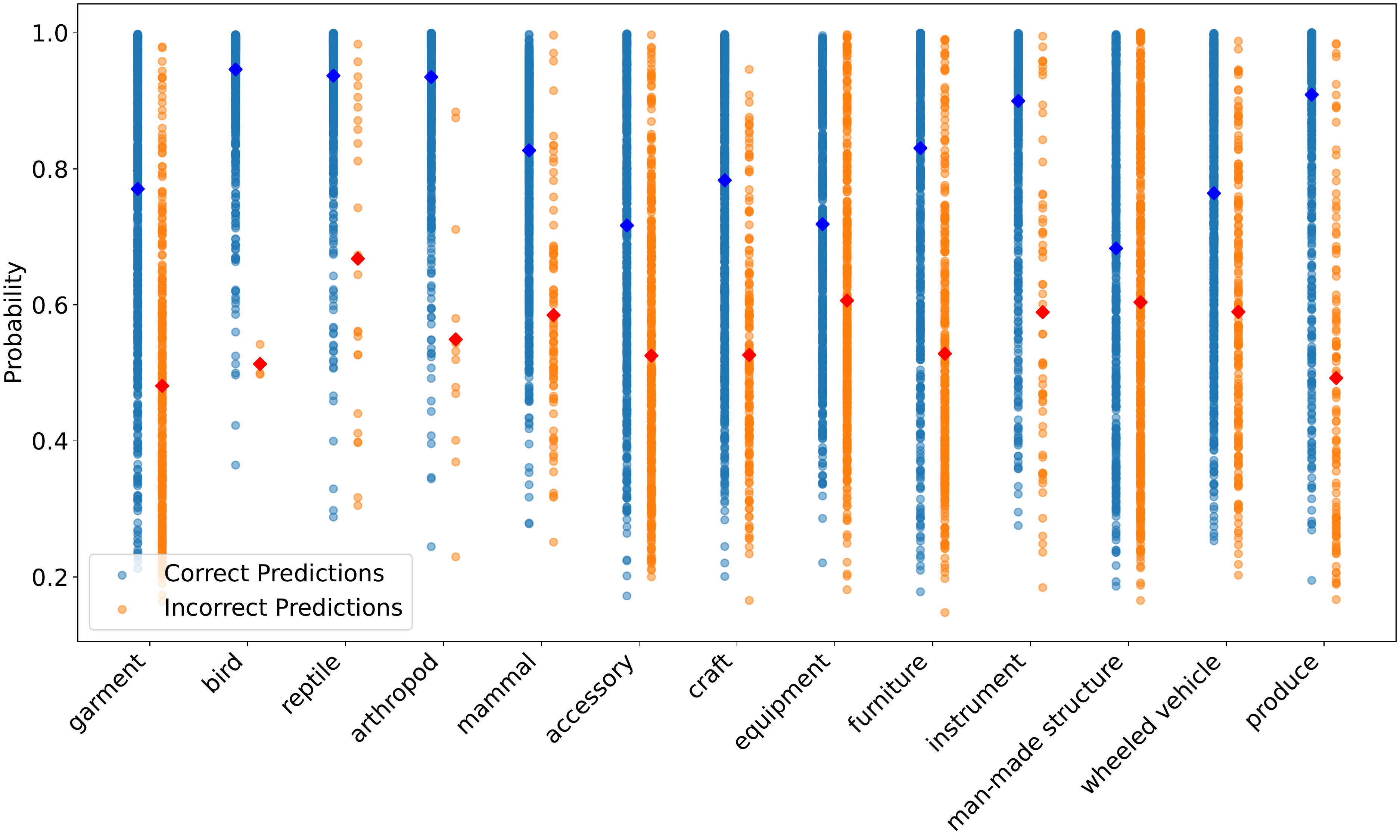}}
     \subfigure[entity30]{\label{fig:e30conf}\includegraphics[width=0.8\textwidth]{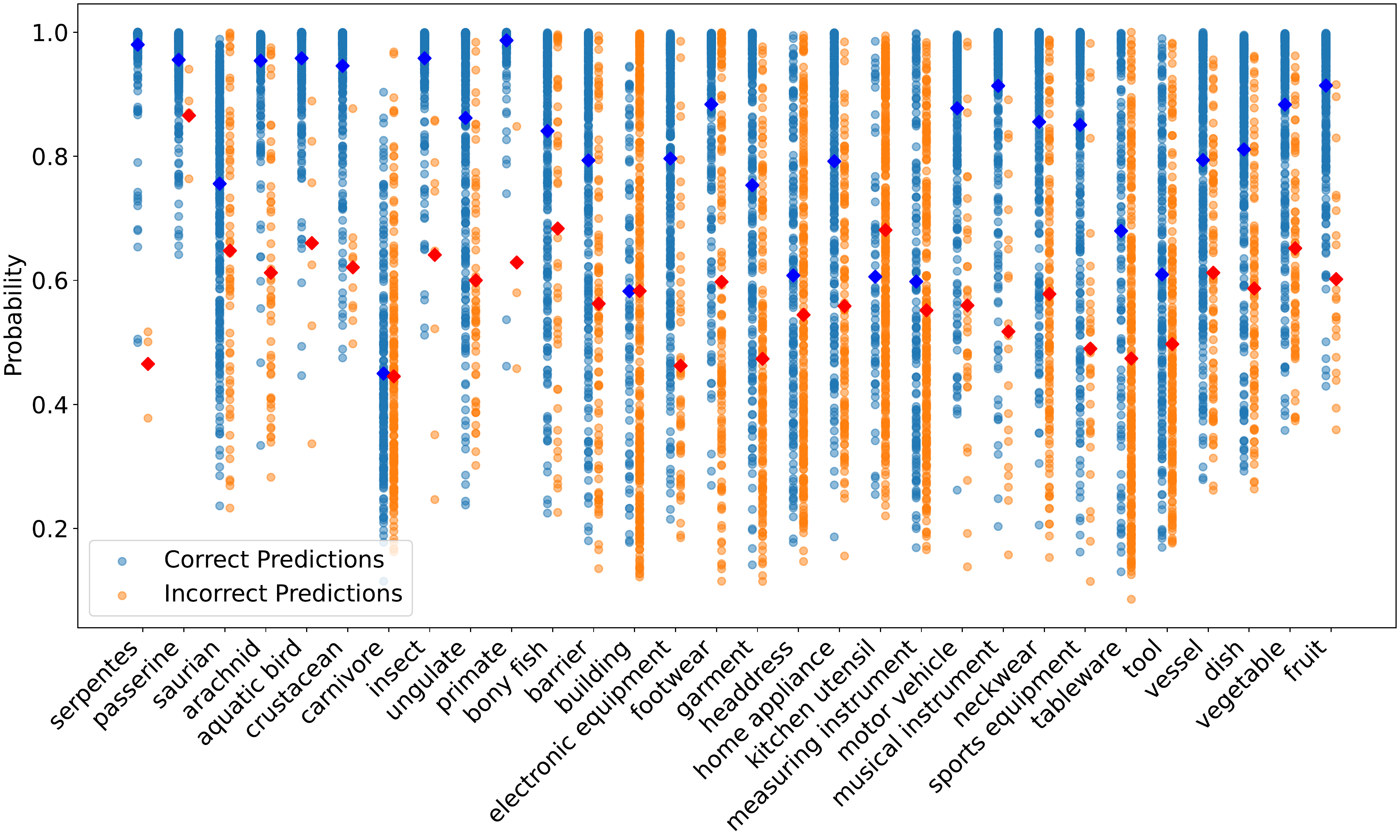}}
     
\end{figure}

The motivation behind the reweighting 
step of \acro{} primarily comes
from the heuristic that LLMs make correct predictions
with high estimated probabilities assigned to them \citep{kadavath2022know},
\change{and that CLIP models themselves are well-calibrated \citep{minderer2021revisiting}}. 
However, we also verify
whether there is some evidence of this behavior in CLIP models.
Given that the output of a CLIP model is a probability
distribution over the provided classes, we care specifically
about the probability of the \emph{argmax} class 
(i.e. the predicted class) when the model is correct
and when it is incorrect.
Across the BREEDS datasets for the standard ImageNet
domain, in Figure \ref{fig:confs} we show the distribution
of the correct and incorrect argmax probabilities for each class
(i.e. for each class $c_i$, we show the output probabilties for $c_i$ when it was correctly classified and the output probabilities
of the predicted classes when the true class is $c_i$).
Whenever CLIP is correct, the associated
probability is on average much higher than
the probabilities associated with misclassification.

\section{\change{CLIP Primer}}\label{app:primer}
\change{\emph{Open Vocabulary models} (as termed in \citet{pham2021scaling}) refer
to models that are able to classify images by associating them with natural
language descriptions of each class. These models are ``open" in the sense
that they are to predict on an \emph{arbitrary} vocabulary of descriptions
(as opposed to a fix set), thus allowing for arbitrary-way image classification.
Popular open vocabulary models include the model of focus CLIP \citep{radford2021learning}
and ALIGN \citep{jia2021scaling} as examples.}

\change{\textbf{C}ontrastive \textbf{L}anguage \textbf{I}mage \textbf{P}retraining (CLIP)
is a family of open vocabulary models, and the focus of the present work.
CLIP, which is comprised of a text encoder and an image encoder that project into
the same latent space, is trained in the following way: Given a set of image-caption
pairs (e.g. a photo of a dog with the caption ``a photo of a dog.''), CLIP is trained
to predict which caption goes with which image as a contrastive learning objective by
comparing the similarity between each image embedding and each caption embedding.}

\change{At inference time (in the zero-shot setting), a naïve method for image classification
(which is the initial baseline tried in \citet{radford2021learning}) involves simply
passing in the list of class names for a given dataset, and calculating the similarity
between a particular image embedding and each one of these class embeddings. However,
\citet{radford2021learning} found that by taking a cue from the recent literature
on prompt engineering for large language models \citep{gao2020making}, CLIP can perform
significantly better as a zero-shot predictor if each class name is included in a
natural language \textbf{prompt} that resembles some sort of image caption (as that
is what CLIP was trained on). As an example, the standard baseline prompt mentioned
is \emph{``A photo of a \{\}."}. In our work, we define a prompt (or prompt template, which
we use interchangeably) as any caption-like
phrase in natural language that a class name can be injected into.}

\section{\change{Adding Context to Prompts and GPT-3 Queries}}\label{app:context}

\begin{table}[ht!]
    
    \caption{\change{Context tokens and prompt sets used for each dataset.}}
    \label{tab:prompts}
    \centering
    \begin{small}
        \change{\begin{tabular}{lcc}
    \toprule
     Dataset & \textbf{[context]} & Prompt Set Used\\
     \midrule
    Nonliving26 & N/A & ImageNet\\
    Living17 & N/A & ImageNet\\
    Entity13 & N/A & ImageNet\\
    Entity30 & N/A & ImageNet\\
    CIFAR20 & N/A & ImageNet\\
    Food-101 & ``food" & Dataset-Specific\\
    Fruits-360 & ``fruit" & Dataset-Specific\\
    Fashion1M & ``article of clothing" & Dataset-Specific\\
    Fashion-MNIST & ``article of clothing" & ImageNet\\
    LSUN-Scene & N/A & ImageNet\\
    Office31 & ``office supply" & Dataset-Specific\\
    OfficeHome & ``office supply" & ImageNet\\
    ObjectNet & N/A & ImageNet\\
    EuroSAT & N/A & Dataset-Specific\\
    RESISC45 & N/A & Dataset-Specific\\
    \bottomrule
    \end{tabular}}
    \end{small}

\end{table}

\change{In order to disentangle the effect that well-formed prompt templates
have on the success of \acro{}, for each dataset (besides the BREEDS
datasets and ObjectNet as they are already semantically similar to ImageNet)
we compare the ImageNet 75 classes against a dataset-specific set of
prompt templates. In the case of EuroSAT, RESISC45, CIFAR20 and Food-101,
we directly use the prompt template set from \citet{radford2021learning}.
For LSUN-Scene, we use the prompt template set for SUN397 \citep{sun397},
as the two datasets are semantically similar. For the rest of the datasets
not yet mentioned (namely Fruits360, Fashion1M, Fashion-MNIST, Office31, 
and OfficeHome) we add the \textbf{[context]} marker into the standard
prompt template as mentioned in Section \ref{sec:setup}. The prompt
sets themselves can be directly found in the code implementation
for this project.}

\change{For the GPT-3 Query with additional context, we add the respective
\textbf{[context]} token to the query \emph{if} the dataset-specific
prompt template is used. Note that we did not create \textbf{[context]}
tokens for EuroSAT, LSUN-Scene, or RESISC45 despite
testing dataset-specific prompt templates, as there did not
seem to be a concise semantic label to describe the classes 
in these datasets. In Table \ref{tab:prompts}, we list the dataset,
the \textbf{[context]} token (if applicable), and the final
prompt set used for all the experiments. Here, we found that
while dataset-specific prompts often improved baseline performance,
they were not \emph{gauranteed} to improve performance,
as in both Fasion-MNIST and OfficeHome the general ImageNet
prompt set performed better.}

\section{\change{Including Superclass Labels in Label Sets}}\label{app:supclass}
\begin{table*}[ht]
    \caption{\change{Zero-Shot Accuracy Performance across benchmarks, controlling for the presence of the superclass label within each respective label set. In the existing map case, adding the superclass labels removes some of the performance gains of the raw existing map. In the GPT-3 Map case, adding the superclass is crucial to maintaining performance in most datasets}}\label{tab:supres}
    \centering
    \footnotesize
    \change{\begin{tabular}{lcccc}
    \toprule
    \multirow{2}{*}{Dataset} & \acro{} Accuracy & \acro{} Accuracy & \acro{} Accuracy & \acro{} Accuracy\\
      &  (Existing Map) & (Existing Map+)& (GPT-3 Map) & (GPT-3 Map+)\\
    \midrule
    Nonliving26  & 90.68 \textcolor{good}{(+10.85)} & 89.80 \textcolor{good}{(+9.97)} & 81.46 \textcolor{good}{(+1.63)} & 81.68 \textcolor{good}{(+1.85)} \\
    Living17 & 93.81 \textcolor{good}{(+2.72)} & 93.62 \textcolor{good}{(+2.53)} & 91.30 \textcolor{good}{(+0.21)} & 91.56 \textcolor{good}{(+0.46)} \\
    Entity13  & 92.59 \textcolor{good}{(+15.13)} & 92.06 \textcolor{good}{(+14.60)} & 76.97 \textcolor{bad}{(-0.48)} & 78.10 \textcolor{good}{(+0.65)} \\
    Entity30  & 88.87 \textcolor{good}{(+18.55)} & 87.29 \textcolor{good}{(+16.96)} & 71.79 \textcolor{good}{(+1.47)} & 71.72 \textcolor{good}{(+1.39)} \\
    CIFAR20  & 85.28 \textcolor{good}{(+25.71)} & 81.45 \textcolor{good}{(+21.88)} & 65.67 \textcolor{good}{(+6.10)} & 65.05 \textcolor{good}{(+5.48)} \\
    Food-101 & N/A & N/A & 93.66 \textcolor{bad}{(-0.21)} & 93.82 \textcolor{bad}{(-0.05)}\\
    Fruits-360  & 59.22 \textcolor{good}{(+0.48)} & 58.88 \textcolor{good}{(+0.15)} & 60.53 \textcolor{good}{(+1.79)} & 60.14 \textcolor{good}{(+1.40)} \\
    Fashion1M  & N/A & N/A & \change{47.51}  \textcolor{good}{($+1.73 $)}& \change{47.44}  \textcolor{good}{($+1.66 $)}\\
    Fashion-MNIST  & N/A & N/A & 70.79 \textcolor{good}{(+2.27)} & 70.81 \textcolor{good}{(+2.29)}\\
    LSUN-Scene  & N/A & N/A & 88.80 \textcolor{good}{($+0.67 $)}& 88.83  \textcolor{good}{($+0.70 $)}\\
    Office31  & N/A & N/A & 86.58 \textcolor{bad}{($-2.71 $)} & \change{90.55}  \textcolor{good}{($+1.42 $)}\\
    OfficeHome  & N/A & N/A & 87.88 \textcolor{bad}{($-0.97 $)}& 88.76  \textcolor{bad}{($-0.09$)}\\
    ObjectNet  & 85.34  \textcolor{good}{($+32.24 $)} & 81.30  \textcolor{good}{($+28.20 $)}& 51.23  \textcolor{bad}{($-2.07$)} & 53.53  \textcolor{good}{($+0.41 $)} \\
    \change{EuroSAT} & \change{N/A} & N/A & \change{62.21} \textcolor{good}{($+0.10$)}& \change{62.40} \textcolor{good}{($+0.29$)} \\
    \change{RESISC45} & N/A & N/A & 71.84 \textcolor{bad}{(-0.75)} & 72.71 \textcolor{good}{(+0.12)}\\
    \bottomrule
    \end{tabular}}
     \vspace{5pt}
\end{table*}

\change{With \acro{} when the existing map is not
available, we append the superclass name to
each label set to account for possible noise
in the GPT-generated label set. In Table \ref{tab:supres},
we show the effect that this inclusion has in both the
existing map and GPT-map cases. Note that in the main paper,
columns 1 and 4 correspond to the main results (i.e. no
superclass labels in existing maps and superclass labels in
GPT-3 maps). In both cases, the presence of the superclass
label more effectively strikes a balance between subclass
and superclass predictions. In the existing map case, this
actually \emph{hurts} performance, as the subclass labels
are optimal in the given dataset. In the GPT-3 map case,
while there are some datasets where removing the superclass
label improves performance (namely Fruits360 and Entity30),
in ever other case removing the superclass label hurts performance,
sometimes by multiple percentage points.}

\section{Label Set Ablation Accuracy}\label{app:lsaa}

\begin{table*}[ht]
    \caption{Accuracy across different label set sizes generated by GPT-3, \change{with best performing label set size in each row bolded}. In general, there is no consistent trend related to label set size and zero-shot performance across datasets.}
    \label{tab:lsaa}
    \centering
    \footnotesize
    \begin{tabular}{lccccc}
    \toprule
     \multirow{2}{*}{Dataset} & \acro{} & \acro{}& \acro{} & \acro{} & \acro{}\\
    & ($m = 1$) &  ($m = 5$) &  ($m = 10$) & ($m = 15$) & ($m = 50$)\\
     \midrule
    Nonliving26 & 79.71 \textcolor{bad}{($-0.12$)} & 81.12 \textcolor{good}{($+1.29$)} & 81.68 \textcolor{good}{($+1.85$)} & \textbf{81.98} \textcolor{good}{($+2.15$)} & 80.03 \textcolor{good}{($+0.20$)} \\
Living17 & 91.14 \textcolor{good}{($+0.04$)} & \textbf{92.68} \textcolor{good}{($+1.58$)} & 91.56 \textcolor{good}{($+0.46$)} & 91.73 \textcolor{good}{($+0.63$)} & 91.41 \textcolor{good}{($+0.31$)} \\
Entity13 & 77.43 \textcolor{bad}{($-0.02$)} & 78.14 \textcolor{good}{($+0.69$)} & 78.10 \textcolor{good}{($+0.65$)} & \textbf{78.37} \textcolor{good}{($+0.92$)} & 78.28 \textcolor{good}{($+0.83$)} \\
Entity30 & 71.06 \textcolor{good}{($+0.73$)} & 71.48 \textcolor{good}{($+1.15$)} & 71.72 \textcolor{good}{($+1.39$)} & \textbf{73.03} \textcolor{good}{($+2.70$)} & 72.62 \textcolor{good}{($+2.29$)} \\
CIFAR20 & 60.15 \textcolor{good}{($+0.58$)} & 64.93 \textcolor{good}{($+5.36$)} & \textbf{65.05} \textcolor{good}{($+5.48$)} & 63.71 \textcolor{good}{($+4.14$)} & 64.99 \textcolor{good}{($+5.42$)} \\
Food-101 & 93.84 \textcolor{bad}{($-0.03$)} & \textbf{93.90} \textcolor{good}{($+0.03$)} & 93.82 \textcolor{bad}{($-0.05$)} & 93.81 \textcolor{bad}{($-0.06$)} & 93.73 \textcolor{bad}{($-0.14$)} \\
Fruits360 & 58.70 \textcolor{bad}{($-0.04$)} & 59.70 \textcolor{good}{($+0.96$)} & \textbf{60.14} \textcolor{good}{($+1.40$)} & 59.75 \textcolor{good}{($+1.01$)} & 59.66 \textcolor{good}{($+0.92$)} \\
Fashion1M & 43.46 \textcolor{bad}{($-2.32$)} & 45.77 \textcolor{bad}{($-0.01$)} & \textbf{47.44} \textcolor{good}{($+1.66$)} & 46.95 \textcolor{good}{($+1.17$)} & 43.61 \textcolor{bad}{($-2.17$)} \\
Fashion-MNIST & 68.01 \textcolor{bad}{($-0.51$)} & \textbf{71.00} \textcolor{good}{($+2.48$)} & 70.81 \textcolor{good}{($+2.29$)} & 69.07 \textcolor{good}{($+0.55$)} & 69.45 \textcolor{good}{($+0.93$)} \\
LSUN-scene & 88.43 \textcolor{good}{($+0.30$)} & 86.30 \textcolor{bad}{($-1.83$)} & \textbf{88.83} \textcolor{good}{($+0.70$)} & 86.80 \textcolor{bad}{($-1.33$)} & 85.97 \textcolor{bad}{($-2.16$)} \\
Office31 & 89.51 \textcolor{good}{($+0.38$)} & 88.15 \textcolor{bad}{($-0.98$)} & \textbf{90.55} \textcolor{good}{($+1.42$)} & 89.43 \textcolor{good}{($+0.30$)} & 89.42 \textcolor{good}{($+0.29$)} \\
OfficeHome & 88.75 \textcolor{bad}{($-0.12$)} & 89.11 \textcolor{good}{($+0.24$)} & 88.76 \textcolor{bad}{($-0.09$)} & \textbf{89.16} \textcolor{good}{($+0.29$)} & 88.87 \textcolor{good}{($+0.00$)} \\
ObjectNet & 53.75 \textcolor{good}{($+0.63$)} & 53.27 \textcolor{good}{($+0.15$)} & 53.53 \textcolor{good}{($+0.41$)} & 57.70 \textcolor{good}{($+4.58$)} & \textbf{58.03} \textcolor{good}{($+4.91$)} \\
EuroSAT & 62.32 \textcolor{good}{($+0.21$)} & 62.21 \textcolor{good}{($+0.10$)} & 62.40 \textcolor{good}{($+0.29$)} & \textbf{62.72} \textcolor{good}{($+0.61$)} & 62.11 \textcolor{good}{($0.00$)} \\
RESISC45 & \textbf{73.29} \textcolor{good}{($+0.70$)} & 73.05 \textcolor{good}{($+0.46$)} & 72.71 \textcolor{good}{($+0.12$)} & 72.67 \textcolor{good}{($+0.08$)} & 71.90 \textcolor{bad}{($-0.69$)} \\
    \bottomrule
    \end{tabular}
\end{table*}

Table \ref{tab:lsaa} displays the raw accuracy 
scores for \acro{} across different label set
sizes.

\section{\change{Alternative Aggregation Methods (Cont.)}}\label{app:aer}

\begin{table}[ht]
\caption{Average accuracy across datasets for varying \change{aggregative} methods on both the prompt and subclass steps of the zero-shot pipeline. In general, linear averaging for subclasses performs worse than our proposed set-based method, while linear averaging for prompts (for raw superclass prediction) performs better thant using a set-based mapping.}\label{tab:ens}
    \centering
    \begin{small}
        \begin{tabular}{lc}
        \toprule
          Experiment  &  Accuracy \\
          \midrule
          Superclass (linear average) & 73.28 \\
          Superclass (set-based prompt mapping) & 72.25 \\
          \acro{} (True Map, set-based mapping) & 85.11 \\
          \acro{} (True Map, linear average) & 81.61 \\
          \acro{} (GPT Map, set-based mapping) & 74.43 \\
          \acro{} (GPT Map, linear average) & 72.25 \\
          \bottomrule
        \end{tabular}
    \end{small}

\end{table}

While \acro{} is based on a \emph{set-based} \change{mapping}
approach for subclasses and a linear averaging
for prompt templates (based on \citet{radford2021learning}'s
procedure), we experimented with \change{two} alternative 
ensembling methods for different parts of the \acro{} pipeline: \change{(1) Using a \emph{linear average} of subclass embeddings rather
    than the set-based mapping (that is, every superclass's text embedding is the
    average across all subclass embeddings, each themselves averaged across every prompt template) and 
    (2) Using a \emph{set-based} mapping for prompt templates rather
    than a linear average (i.e. instead of averaging across prompt templates, predict across each prompt template separately at inference time and then use embedded class to map back to the set of superclasses).}
Note in the latter case we only experiment with how this
effects \emph{superclass} prediction (where each class
maps to a set of \change{the dataset's chosen} prompt embeddings), as
using set-based ensembling for \emph{both} prompts
and subclasses within \acro{} quickly becomes 
computationally expensive. In Table \ref{tab:ens},
we see that using our initial \change{aggregation} methods
(i.e. linear averaging for prompts and set mappings
for subclasses) achieves greater accuracy.

\section{\change{Noisy Available Hierarchy Details}}\label{app:noise}
\change{The ImageNet \citep{deng2009imagenet} dataset itself
includes a rich hierarchical taxonomy, where every class is a leaf
node of the hierarchy. 
In the original BREEDS \citep{santurkar2020breeds} work,
the authors modify the structure slightly in order to place
concepts at semantically-similar levels of granularity at the same depth,
and additional restrict the number of subclasses 
within each of the BREEDS datasets in order to balance the data.
Thus, it is possible for each BREEDS dataset to use the dataset
with its superclasses and restricted set of subclasses
but provide \acro{} with \emph{all} the subclass labels present in the ImageNet
hierarchy for each superclass (i.e. all leaf nodes descended from each superclass node). 
In Table \ref{tab:nah_ex}, we display
a subset of the living17 BREEDS dataset class structure with the original
subclasses and the ImageNet subclasses. Observe that in some cases, 
there are many subclass labels provided to \acro{} than is present in the data.}

\section{Dataset Details}\label{app:dataset}

\begin{table}[ht]
\caption{Domains used for BREEDS, Office31, and OfficeHome.}
    \label{tab:domains}
    \centering
    \begin{tabular}{cc}
    \toprule
    Dataset & Domains \\
    \midrule
    \multirow{2}{*}{BREEDS}    & ImageNet, ImageNet-Sketch, ImageNetv2, ImageNet-c \\ & \{Fog-1, Contrast-2, Snow-3, Gaussian Blur-4, Saturate-5\}\\
    Office31 & Amazon, DSLR, webcam\\
    OfficeHome & Clipart, Art, Real World, Product\\
    \bottomrule
    \end{tabular}
    
\end{table}

\paragraph{\acro{} Across Domain Shifts} For each
of the BREEDS datasets \citep{santurkar2020breeds}, Office31 \citep{office31}, and OfficeHome \citep{officehome}, 
all results presented are the average over different domains.
The specific domains used are show in Table \ref{tab:domains}.

\begin{table*}[t!]
    \caption{\change{Subset of living17 class hierarchy, showing the difference between the original BREEDS subclasses and the ImageNet subclasses used for the ablation in Section \ref{sec:abl}: Noisy Available Hierarchies.}}
    \label{tab:nah_ex}
    \centering
    \footnotesize
    \change{\begin{tabular}{p{1.5cm}p{5cm}p{5cm}}
    \toprule
     Superclass & Original BREEDS subclasses & All ImageNet subclasses\\
     \midrule
     salamander & European fire salamander, common newt, eft, spotted salamander & European fire salamander, common newt, eft, spotted salamander, axolotl\\
turtle & loggerhead, leatherback turtle, mud turtle, terrapin & loggerhead, leatherback turtle, mud turtle, terrapin, box turtle\\
lizard & common iguana, American chameleon, agama, frilled lizard & banded gecko, common iguana, American chameleon, whiptail, agama, frilled lizard, alligator lizard, Gila monster, green lizard, African chameleon, Komodo dragon\\
snake & thunder snake, ringneck snake, diamondback, sidewinder & thunder snake, ringneck snake, hognose snake, green snake, king snake, garter snake, water snake, vine snake, night snake, boa constrictor, rock python, Indian cobra, green mamba, sea snake, horned viper, diamondback, sidewinder\\
spider & black and gold garden spider, barn spider, garden spider, black widow & black and gold garden spider, barn spider, garden spider, black widow, tarantula, wolf spider\\
grouse & black grouse, ptarmigan, ruffed grouse, prairie chicken & black grouse, ptarmigan, ruffed grouse, prairie chicken\\
parrot & African grey, macaw, sulphur-crested cockatoo, lorikeet & African grey, macaw, sulphur-crested cockatoo, lorikeet\\
crab & Dungeness crab, rock crab, fiddler crab, king crab & Dungeness crab, rock crab, fiddler crab, king crab\\
\bottomrule
    \end{tabular}}
\end{table*}

\paragraph{Fruits-360} For zero-shot classification with
CLIP models, Fruits-360 \citep{fruits360} in its raw form is somewhat ill-formed
from a class name perspective, as there are classes only differentiated
by a numeric index (e.g. ``Apple Golden 1" and ``Apple Golden 2") and 
classes at mixed granularity (e.g. ``forest nut" and ``hazelnut" are 
separate classes even though hazelnuts are a type of forest nut).
We thus manually rename classes using the structure laid out in
Table \ref{tab:fruits}, which results in a 59-way superclass
classification problem, with 102 ground-truth subclasses.

\subsection{ObjectNet: A Case Study}

The ObjectNet dataset \citep{objectnet} has partial overlap (113 classes) with the ImageNet \citep{deng2009imagenet} hierarchical class structure.
From this subset of ObjectNet, we use the BREEDS hierarchy
\citep{santurkar2020breeds} to generate a coarse-grained
version of ObjectNet that is shown in Table \ref{tab:object}.
In this 11-way classification task, the true subclasses are
the original ObjectNet classes.
Additionally, here we show the GPT-generated subsets at $m=10$.

In observing the ground truth vs. generated subsets for each class in ObjectNet, we can see that for the most part, GPT-3 fails to accurately guess most of the true subclasses, even in the case when the true number of subclasses is quite small. This is quite noticeable in classes such as ``equipment" and ``cooked food", where GPT-3 gets \emph{none} of the subclasses correct. Thus, we posit that this behavior is the root cause for the relative poor performance of \acro{} when using GPT-generated subsets, as here in ObjectNet (and more broadly) superclass names may not be great indicators for the true subclass distribution. In ObjectNet in particular, the relative ambiguity of class names like ``accessory," ``appliance," and ``equipment" most likely contribute to the poor baseline performance, as well as ObjectNet's inherent difficulty by design.

\begin{table*}[ht]
    \caption{Class Structure for ObjectNet experiments.}
    \label{tab:object}
    \centering
    \begin{small}
            \begin{tabular}{p{2.5cm}p{5cm}p{5cm}}
    \toprule
    Superclass & Subclasses (Original ObjectNet) & Subclasses (GPT-Generated)\\
    \midrule
    garment & Dress, Jeans, Skirt, Suit jacket, Sweater, Swimming trunks, T-shirt & T-shirt, dress, skirt, blouse, pants, shorts, leggings, jeans, overalls, jumpsuit \\
soft furnishings & Bath towel, Desk lamp, Dishrag or hand towel, Doormat, Lampshade, Paper towel, Pillow & curtains, drapes, blinds, shades, valances, swags, cornices, drapery hardware, upholstery, slipcovers \\
accessory & Backpack, Dress shoe (men), Helmet, Necklace, Plastic bag, Running shoe, Sandal, Sock, Sunglasses, Tie, Umbrella, Winter glove & earrings, necklace, bracelet, ring, brooch, belt, scarf, gloves, hat, glasses\\
appliance & Coffee/French press, Fan, Hair dryer, Iron (for clothes),  Microwave, Portable heater, Toaster, Vacuum cleaner & blender, coffee maker, toaster, mixer, crock pot, rice cooker, dishwasher, dryer, washer, oven\\
equipment & Cellphone, Computer mouse, Keyboard, Laptop (open), Monitor, Printer, Remote control, Speaker, Still Camera, TV, Tennis racket, Weight (exercise) & trowel, hoe, rake, shovel, bucket, wheelbarrow, watering can, shears, gloves, hat\\
furniture & Bench, Chair & table, chair, dresser, bed, nightstand, lamp, couch, loveseat, coffee table, end table\\
toiletry & Band Aid, Lipstick & toothbrush, toothpaste, floss, mouthwash, soap, shampoo, conditioner, body wash, lotion, deodorant\\
wheeled vehicle & Basket, Bicycle & car, bus, train, bike, skateboard, rollerblades, wheelchair, tractor, dune buggy, gokart\\
cooked food & Bread loaf & stir fry, spaghetti, soup, salad, roast, rice, quinoa, pancakes, omelette, pasta\\
produce & Banana, Lemon, Orange & apple, banana, orange, grapefruit, lemon, lime, watermelon, cantaloupe, honeydew, pineapple\\
beverage & Drinking Cup & coffee, tea, water, soda, milk, orange juice, apple juice, grape juice, cranberry juice, tomato juice\\
\bottomrule
    \end{tabular}
    \end{small}

\end{table*}

\begin{longtable}[c]{ccc}
\caption{Mapping from original class names to new subclass and superclasses for Fruits-360.}\label{tab:fruits}\\
    \toprule
    Original Class & Cleaned Subclass & Cleaned Superclass \\
    \midrule
     Apple Braeburn & braeburn apple & apple \\
Apple Crimson Snow & crimson snow apple & apple \\
Apple Golden 1 & golden apple & apple \\
Apple Golden 2 & golden apple & apple \\
Apple Golden 3 & golden apple & apple \\
Apple Granny Smith & granny smith apple & apple \\
Apple Pink Lady & pink lady apple & apple \\
Apple Red 1 & red apple & apple \\
Apple Red 2 & red apple & apple \\
Apple Red 3 & red apple & apple \\
Apple Red Delicious & red delicious apple & apple \\
Apple Red Yellow 1 & red yellow apple & apple \\
Apple Red Yellow 2 & red yellow apple & apple \\
Apricot & apricot & apricot \\
Avocado & avocado & avocado \\
Avocado ripe & avocado & avocado \\
Banana & banana & banana \\ 
 \endfirsthead
Banana Lady Finger & lady finger banana & banana \\
Banana Red & red banana & banana \\
Beetroot & beetroot & beetroot \\
Blueberry & blueberry & blueberry \\
Cactus fruit & cactus fruit & cactus fruit \\
Cantaloupe 1 & melon & melon \\
Cantaloupe 2 & melon & melon \\
Carambula & star fruit & star fruit \\
Cauliflower & cauliflower & cauliflower \\
Cherry 1 & cherry & cherry \\
Cherry 2 & cherry & cherry \\
Cherry Rainier & rainier cherry & cherry \\
Cherry Wax Black & black cherry & cherry \\
Cherry Wax Red & red cherry & cherry \\
Cherry Wax Yellow & yellow cherry & cherry \\
Chestnut & nut & nut \\
Clementine & orange & orange \\
Cocos & cocos & cocos \\
Corn & corn & corn \\
Corn Husk & corn husk & corn husk \\
Cucumber Ripe & cucumber & cucumber \\
Cucumber Ripe 2 & cucumber & cucumber \\
Dates & date & date \\
Eggplant & eggplant & eggplant \\
Fig & fig & fig \\
Ginger Root & ginger root & ginger root \\
Granadilla & granadilla & passion fruit \\
Grape Blue & blue grape & grape \\
Grape Pink & pink grape & grape \\
Grape White & white grape & grape \\
Grape White 2 & white grape & grape \\
Grape White 3 & white grape & grape \\
Grape White 4 & white grape & grape \\
Grapefruit Pink & pink grapefruit & grapefruit \\
Grapefruit White & white grapefruit & grapefruit \\
Guava & gauva & gauva \\
Hazelnut & nut & nut \\
Huckleberry & huckleberry & huckleberry \\
Kaki & kaki & persimmon \\
Kiwi & kiwi & kiwi \\
Kohlrabi & kohlrabi & kohlrabi \\
Kumquats & kumquat & kumquat \\
Lemon & lemon & lemon \\
Lemon Meyer & meyer lemon & lemon \\
Limes & lime & lime \\
Lychee & lychee & lychee \\
Mandarine & orange & orange \\
Mango & mango & mango \\
Mango Red & red mango & mango \\
Mangostan & mangostan & mangostan \\
Maracuja & maracuja & passion fruit \\
Melon Piel de Sapo & melon & melon \\
Mulberry & mulberry & mulberry \\
Nectarine & nectarine & nectarine \\
Nectarine Flat & flat nectarine & nectarine \\
Nut Forest & forest nut & nut \\
Nut Pecan & pecan nut & nut \\
Onion Red & red onion & onion \\
Onion Red Peeled & red onion & onion \\
Onion White & white onion & onion \\
Orange & orange & orange \\
Papaya & papaya & papaya \\
Passion Fruit & passion fruit & passion fruit \\
Peach & peach & peach \\
Peach 2 & peach & peach \\
Peach Flat & flat peach & peach \\
Pear & pear & pear \\
Pear 2 & pear & pear \\
Pear Abate & abate pear & pear \\
Pear Forelle & forelle pear & pear \\
Pear Kaiser & kaiser pear & pear \\
Pear Monster & monster pear & pear \\
Pear Red & red pear & pear \\
Pear Stone & stone pear & pear \\
Pear Williams & williams pear & pear \\
Pepino & pepino & pepino \\
Pepper Green & green pepper & pepper \\
Pepper Orange & orange pepper & pepper \\
Pepper Red & red pepper & pepper \\
Pepper Yellow & yellow pepper & pepper \\
Physalis & groundcherry & groundcherry \\
Physalis with Husk & groundcherry & groundcherry \\
Pineapple & pineapple & pineapple \\
Pineapple Mini & mini pineapple & pineapple \\
Pitahaya Red & dragon fruit & dragon fruit \\
Plum & plum & plum \\
Plum 2 & plum & plum \\
Plum 3 & plum & plum \\
Pomegranate & pomegranate & pomegranate \\
Pomelo Sweetie & pomelo & pomelo \\
Potato Red & red potato & potato \\
Potato Red Washed & red potato & potato \\
Potato Sweet & sweet potato & potato \\
Potato White & white potato & potato \\
Quince & quince & quince \\
Rambutan & rambutan & rambutan \\
Raspberry & raspberry & raspberry \\
Redcurrant & redcurrant & redcurrant \\
Salak & salak & snake fruit \\
Strawberry & strawberry & strawberry \\
Strawberry Wedge & strawberry & strawberry \\
Tamarillo & tamarillo & tamarillo \\
Tangelo & tangelo & tangelo \\
Tomato 1 & tomato & tomato \\
Tomato 2 & tomato & tomato \\
Tomato 3 & tomato & tomato \\
Tomato 4 & tomato & tomato \\
Tomato Cherry Red & cherry tomato & tomato \\
Tomato Heart & heart tomato & tomato \\
Tomato Maroon & maroon tomato & tomato \\
Tomato Yellow & yellow tomato & tomato \\
Tomato not Ripened & unripe tomato & tomato \\
Walnut & nut & nut \\
Watermelon & melon & melon \\
\bottomrule
    \label{tab:my_label}
\end{longtable}

\end{document}